\documentclass[10pt,twocolumn,letterpaper]{article}

\usepackage[pagenumbers]{cvpr} 
\usepackage{pifont}
\usepackage{xcolor}
\usepackage[table]{xcolor}
\usepackage{multirow}

\usepackage{tikz}
\usepackage{makecell}
\usepackage{booktabs} 

\usepackage{listings}
\usepackage{xcolor}
\usepackage{placeins}

\lstdefinelanguage{yaml}{
  keywords={true,false,null},
  morestring=[b]',
  morestring=[b]",
  sensitive=false,
  morecomment=[l]{\#},
}

\definecolor{cvprblue}{rgb}{0.21,0.49,0.74}
\usepackage[pagebackref,breaklinks,colorlinks,allcolors=cvprblue]{hyperref}

\title{DriveCombo: Benchmarking Compositional Traffic Rule Reasoning\\ in Autonomous Driving}

\author{
    Enhui Ma\textsuperscript{1}\footnotemark[1]~\footnotemark[2]~,
    Jiahuan Zhang\textsuperscript{1}\footnotemark[1]~\footnotemark[2]~,
    Guantian Zheng\textsuperscript{1}\footnotemark[3]~,
    Tao Tang\textsuperscript{2}\footnotemark[2]~,
    Shengbo Eben Li\textsuperscript{3},~
    Yuhang Lu\textsuperscript{4},\\
    Xia Zhou\textsuperscript{2},
    Xueyang Zhang\textsuperscript{2},
    Yifei Zhan\textsuperscript{2}, 
    Kun Zhan\textsuperscript{2},
    Zhihui Hao\textsuperscript{2},
    Xianpeng Lang\textsuperscript{2},
    Kaicheng Yu\textsuperscript{1}\footnotemark[4] \\
    \textsuperscript{1}Autolab, Westlake University~
    \textsuperscript{2}Li Auto Inc~
    \textsuperscript{3}Tsinghua University~
    \textsuperscript{4}The University of Hong Kong\\
    \begin{normalsize}${\tt \{maenhui, kyu\}@westlake.edu.cn}$\end{normalsize}
}

\begin{document}

\twocolumn[{%
\renewcommand\twocolumn[1][]{#1}%
\maketitle
\begin{center}
    \vspace{-0.15cm}
    \includegraphics[width=\linewidth]{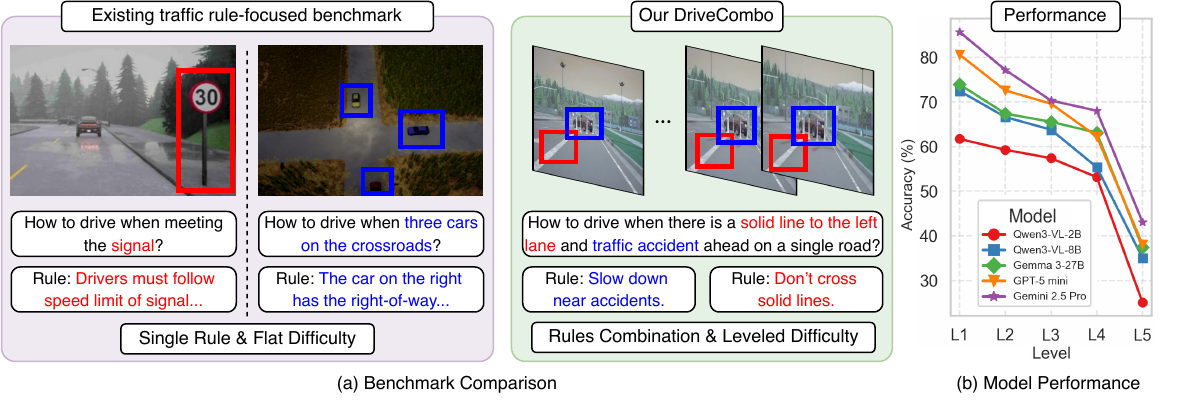}
    \captionof{figure}{
    \textbf{(a)} Existing traffic rule-focused benchmarks~\cite{wei2025driveqa, lu2025idkb} mainly assess single-rule understanding, such as recognizing traffic signs or simple right-of-way cases, resulting in flat difficulty and limited rule reasoning. In contrast, our DriveCombo introduces compositional traffic rule scenarios with leveled cognitive difficulty, enabling systematic evaluation of multimodal large language models (MLLMs) from single-rule understanding to multi-rule integration and conflict resolution.
    \textbf{(b)} Model performance across our Five-Level Cognitive Ladder. For simplicity, we present results for five representative models, which show a consistent decline in accuracy as reasoning complexity increases, especially at Level 5 conflict-resolution task, revealing current limitations in compositional traffic rule reasoning of MLLMs.
}
    \label{fig:teaser}
\end{center}
}]

\renewcommand{\thefootnote}{\fnsymbol{footnote}}
\footnotetext[1]{Co-first authors.}
\footnotetext[2]{Work done during an internship at Li Auto Inc.}
\footnotetext[3]{Work done during their visiting at Autolab, Westlake University.}
\footnotetext[4]{Corresponding Author.}
\renewcommand{\thefootnote}{\arabic{footnote}}

\begin{abstract}
Multimodal Large Language Models (MLLMs) are rapidly becoming the intelligence brain of end-to-end autonomous driving systems.
 A key challenge is to assess whether MLLMs can truly understand and follow complex real-world traffic rules. However, existing benchmarks mainly focus on single-rule scenarios like traffic sign recognition, neglecting the complexity of multi-rule concurrency and conflicts in real driving. Consequently, models perform well on simple tasks but often fail or violate rules in real world complex situations. To bridge this gap, we propose \textbf{DriveCombo}, a text and vision-based benchmark for compositional traffic rule reasoning. Inspired by human drivers’ cognitive development, we propose a systematic \textbf{Five-Level Cognitive Ladder} that evaluates reasoning from single-rule understanding to multi-rule integration and conflict resolution, enabling quantitative assessment across cognitive stages. 
 We further propose a \textbf{Rule2Scene Agent} that maps language-based traffic rules to dynamic driving scenes through rule crafting and scene generation, enabling scene-level traffic rule visual reasoning.
 Evaluations of 14 mainstream MLLMs reveal performance drops as task complexity grows, particularly during rule conflicts. After splitting the dataset and fine-tuning on the training set, we further observe substantial improvements in both traffic rule reasoning and downstream planning capabilities. These results highlight the effectiveness of DriveCombo in advancing compliant and intelligent autonomous driving systems.
\end{abstract}

\section{Introduction}
\label{sec:intro}


With the rapid advancement of multimodal large language models (MLLMs)\cite{openai2025gpt5, yang2025qwen3,team2023gemini}, many efforts have been devoted to harnessing them to boost end-to-end (E2E) autonomous driving systems. 
By fusing visual perception, linguistic reasoning, and world knowledge, they greatly enhance scene understanding and trajectory planning~\cite{xu2024drivegpt4, tian2024drivevlm,cui2025drivemlm, pan2024vlp, hwang2024emma}. However, safe driving depends not only on trajectory safety but also on compliance with traffic regulations, meaning the ability to make lawful and contextually appropriate decisions. However, traditional E2E benchmarks mainly measure planning metrics such as trajectory deviation or collision rate, while neglecting the model’s understanding of driving rules. As a result, even models that plan well may still hesitate, misjudge, or violate regulations in complex scenarios, revealing an urgent need for benchmarks that assess traffic rule reasoning and compliance.

\setlength{\tabcolsep}{6pt}
\begin{table*}[ht!]
\centering
\footnotesize
\caption{\textbf{Comparison of Existing Vision Language Driving Datasets.} ``\#Rules" means the number of traffic rules within each scenario. ``3D Scene" indicates whether image sequence frames information is available.}
\label{tab:benchmark_comparison}
\begin{tabular}{l|ccccccccc}
\toprule
\textbf{Dataset} & \textbf{Venue} & \textbf{Country} & \textbf{QA pairs} & \textbf{Images} & \textbf{Scene source} & \textbf{3D Scene} & \textbf{Traffic Rules} & \textbf{\#Rules} \\
\midrule
Talk2Car~\cite{deruyttere2019talk2car} & EMNLP'19 & 2 & 12K & 1.8K & nuScenes~\cite{caesar2020nuscenes} & {\color{green}\ding{51}} & {\color{red}\ding{55}} & - \\
DRAMA~\cite{malla2023drama} & WACV'23 & JP & 14K & - & DRAMA~\cite{malla2023drama} & {\color{green}\ding{51}} & {\color{red}\ding{55}} & - \\
nuScenes-QA~\cite{qian2024nuscenes-QA} & AAAI'24 & 2 & 450K & 34K & nuScenes~\cite{caesar2020nuscenes} & {\color{green}\ding{51}} & {\color{red}\ding{55}} & - \\
LingoQA~\cite{marcu2024lingoqa} & ECCV'24 & UK & 419K & 28K & LingoQA~\cite{marcu2024lingoqa} & {\color{green}\ding{51}} & {\color{red}\ding{55}} & - \\
VLAAD~\cite{park2024vlaad} & WACV'24 & US & 64K & - & BDD~\cite{yu2020bdd100k} & {\color{green}\ding{51}} & {\color{red}\ding{55}} & - \\
DriveBench~\cite{xie2025DriveBench} & ICCV'25 & 2 & 21K & 19K & nuScenes~\cite{caesar2020nuscenes} & {\color{green}\ding{51}} & {\color{red}\ding{55}} & - \\
CODA-LM~\cite{wang2024coda-lm} & WACV'25 & 3 & 63K & 10K & CODA~\cite{li2022coda} & {\color{red}\ding{55}} & {\color{red}\ding{55}} & - \\ \midrule
IDKB~\cite{lu2025idkb} & AAAI'25 & 15 & 1M & 400K & Web, CARLA~\cite{dosovitskiy2017carla} & {\color{red}\ding{55}} & {\color{green}\ding{51}} & 1 \\
DriveQA~\cite{wei2025driveqa} & ICCV'25 & US & 474K & 68K & CARLA~\cite{dosovitskiy2017carla}, Mapillary~\cite{neuhold2017mapillary} & {\color{red}\ding{55}} & {\color{green}\ding{51}} & 1 \\  \midrule
\textbf{DriveCombo} & - & 5 & 70K  & 280K & CARLA~\cite{dosovitskiy2017carla} & {\color{green}\ding{51}} & {\color{green}\ding{51}} & $\geq$1 \\
\bottomrule
\end{tabular}
\end{table*}

As shown in Table~\ref{tab:benchmark_comparison}, recent benchmarks begin to address the limitations mentioned above. Some~\cite{deruyttere2019talk2car, malla2023drama, qian2024nuscenes-QA, xie2025DriveBench, wang2024coda-lm} extend existing 3D driving datasets~\cite{caesar2020nuscenes, yu2020bdd100k, li2022coda} with visual question answering (QA) tasks to deeply evaluate MLLMs’ visual perception and reasoning capabilities, yet still overlook traffic rule compliance. Concurrently, DriveQA~\cite{wei2025driveqa} and IDKB~\cite{lu2025idkb} have taken early steps toward regulatory evaluation by introducing relevant traffic-rule related QA tasks, as shown in Figure~\ref{fig:teaser} (a). However, they typically focus on single rules and rely on 2D static imagery, such as recognizing traffic signs, rather than reasoning within complex dynamic, rule-interacting environments. In real driving, multiple regulations often coexist or conflict. Thus, these benchmarks’ simplified settings create an illusion of performance: models may correctly identify single rules but fail under multi-rule compositions, exposing a gap between present evaluation paradigms and the cognitive requirements of safe and lawful, real-world driving.

To bridge this gap, we propose \textbf{DriveCombo}, a benchmark designed to evaluate MLLMs’ capability in compositional traffic rule reasoning under complex driving scenarios. Inspired by the cognitive development of human drivers, which progresses from understanding single rules to managing multiple constraints and finally to handling conflicts, we construct a \textbf{Five-Level Cognitive Ladder}. This framework advances from \textbf{L1}, understanding single traffic rules, to \textbf{L2}, integrating relevant multiple static constraints; \textbf{L3}, reasoning about dynamic agent interactions; \textbf{L4}, coordinating both static and dynamic rules; and finally \textbf{L5}, resolving rule conflicts to achieve context-aware and legally compliant decision-making. This hierarchical framework systematically disentangles reasoning complexity across cognitive levels, enabling precise analysis of how models evolve from basic rule recognition to conflict resolution.
 Based on this, we further propose a \textbf{Rule2Scene Agent} that transforms traffic rules into 3D driving scenes suitable for visual question answering. The agent first uses a \textit{Rule Crafter} to construct valid hierarchically rule pairs, and then a \textit{Scene Weaver} converts these rule pairs into executable dynamic scenes in the CARLA simulator~\cite{dosovitskiy2017carla}. By generating structured rule representations on the language side and reconstructing their physical semantics on the simulation side, the agent ensures semantic consistency between the traffic rules and the generated scenes, forming a closed loop from rule reasoning to scene execution. Our contributions are summarized as follows:
\begin{enumerate}
    \item We propose DriveCombo, the first multimodal benchmark dedicated to complex traffic-rule reasoning, establishing a systematic Five-Level Cognitive Ladder that overcomes the limitations of single-rule evaluations;
    \item We propose a Rule2Scene Agent that transforms textual traffic rules into dynamic 3D driving scenarios, enabling scene-level traffic rule visual reasoning;
    \item Comprehensive evaluations on 14 MLLMs reveal consistent performance degradation across the Five-Level Cognitive Ladder, most prominently on the L5 conflict-resolution tasks (see Figure~\ref{fig:teaser} (b)), highlighting a bottleneck in traffic-rule reasoning. Moreover, pretraining on DriveCombo significantly improves downstream E2E planning performance, reducing the L2 loss by 17.3\% and thereby validating the effectiveness of our dataset.
\end{enumerate}

\section{Related Work}

\subsection{MLLMs for Autonomous Driving}
Multimodal Large Language Models (MLLMs) are becoming the intelligent core of end-to-end autonomous driving systems, providing stronger semantic understanding, reasoning, and world knowledge~\cite{cui2025drivemlm,xu2024drivegpt4,mao2023agentdriver,tian2024drivevlm,pan2024vlp,hwang2024emma,chen2024driving,cui2024drive,cui2024receive,fu2024drive,mao2023gpt,sha2023languagempc,wen2023dilu,zhou2024embodied}. These approaches enable a unified intelligent architecture across perception, planning, and decision-making tasks. Although these models enhance autonomous driving systems’ cognitive abilities, their training data are largely from general corpora and lack driving knowledge, resulting in failures in complex, rule-governed scenarios. Thus, a benchmark that encodes systematic driving knowledge and evaluates MLLMs’ compliance with complex traffic rules is essential for improving their reliability.


\subsection{Benchmarks for Autonomous Driving}
\paragraph{Real-world Driving Benchmarks} have long served as the foundation for model training and evaluation. Datasets such as KITTI~\cite{geiger2013KITTI}, Waymo~\cite{sun2020Waymo}, Argoverse~\cite{wilson2023argoverse}, and nuScenes~\cite{caesar2020nuscenes} collect multimodal urban traffic perception data that support tasks including detection, segmentation, and trajectory prediction.
However, these evaluations primarily target routine or well-structured driving scenarios and emphasize physical-level performance metrics—such as trajectory deviation or control accuracy, which fail to capture whether the system truly understands and adheres to traffic laws or higher-level driving rules.

\paragraph{Vision Language Driving Benchmarks.}
With the rise of MLLMs, many benchmarks have emerged to evaluate driving scenarios and semantic understanding. BDD-X~\cite{kim2018BDDX}, BDD-OIA~\cite{xu2020BDDOIA}, and LingoQA~\cite{marcu2024lingoqa} benchmark event comprehension via video explanations; Talk2Car~\cite{deruyttere2019talk2car} evaluates language-guided target localization; nuScenes-QA~\cite{qian2024nuscenes-QA} tests 3D relational reasoning; DriveLM~\cite{sima2023drivelm} addresses multi-stage reasoning across perception, prediction, and planning; VLAAD~\cite{park2024vlaad} probes complex reasoning; and DriveBench~\cite{xie2025DriveBench}, OmniDrive~\cite{wang2025omnidrive}, and SimLingo~\cite{renz2025simlingo} extend evaluation to 3D reasoning and language–action consistency. However, these still focus mainly on perceptual and semantic aspects, lacking systematic assessment of traffic rules and decision-making logic. Consequently, models may show strong visual–semantic understanding yet still fail to capture legally compliant driving behavior.

\paragraph{Traffic Rule-focused Benchmarks.}
Recent efforts such as DriveQA~\cite{wei2025driveqa} and IDKB~\cite{lu2025idkb} investigate models’ understanding of traffic rules by constructing rule-based QA datasets from driving manuals and exam banks and by generating scenarios with CARLA~\cite{dosovitskiy2017carla}. These benchmarks represent early attempts to model driving knowledge at the language level, but their designs remain simplified, covering only single atomic rules and relying on 2D static images, which limits their ability to capture real-world scenarios involving multiple overlapping regulations.

To address these limitations above, we propose DriveCombo, a benchmark for multi-rule compositional reasoning. It systematically evaluates the cognitive hierarchy and reasoning capability of MLLMs in complex regulatory contexts and provides a unified standard for assessing the regulatory compliance of autonomous driving systems.

\begin{figure*}[t!]
    \centering
    \vspace{-0.5cm}
      \includegraphics[width=1.0\linewidth]{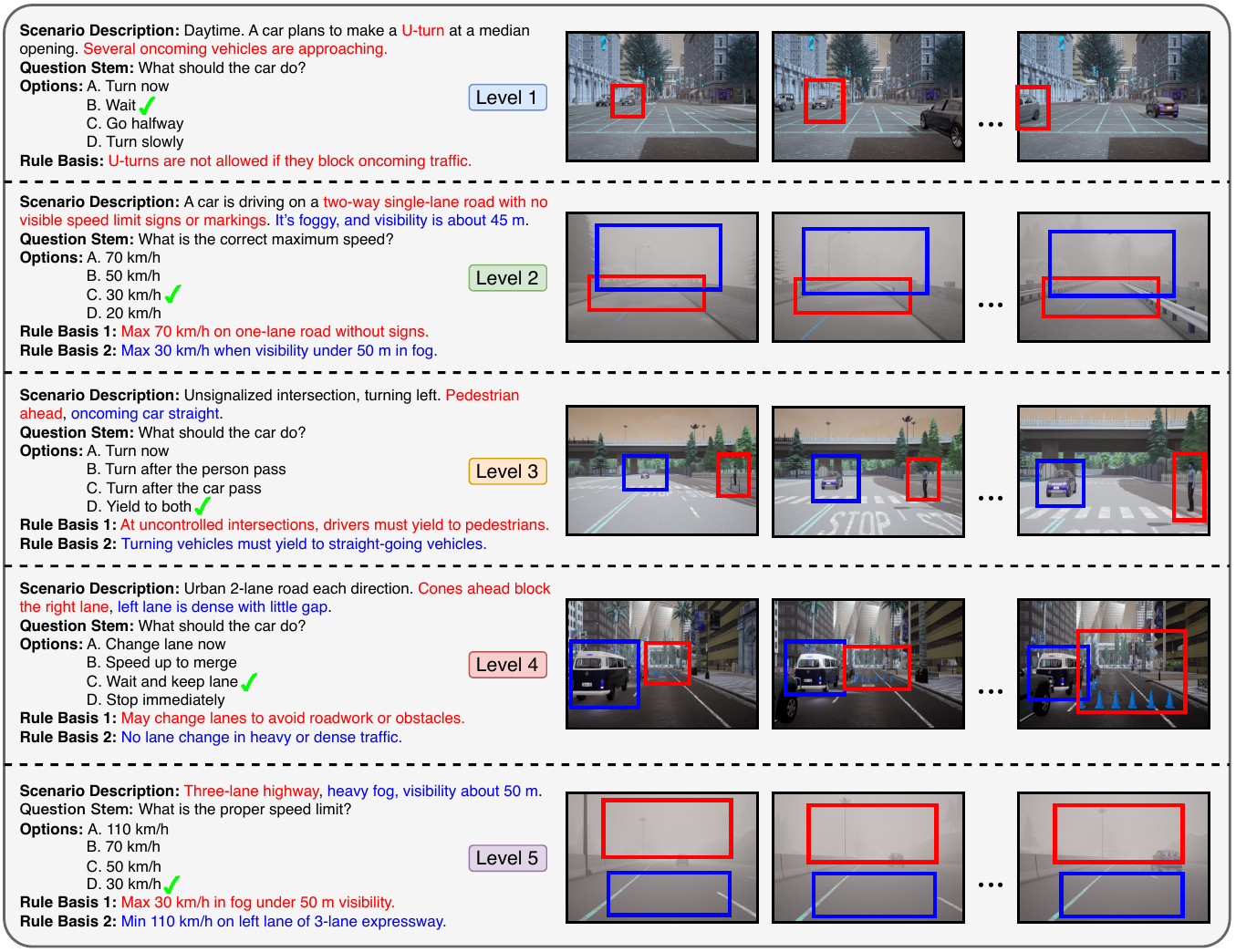}
    \caption{ \textbf{Task Examples of the Five-Level Cognitive Ladder in DriveCombo}, progressing from single-rule understanding (L1) to multi-rule integration and conflict resolution (L5). Each level presents a scenario with corresponding traffic rules bases, visual context, and multiple choice questions, enabling systematic evaluation of MLLMs’ compositional traffic rule reasoning under increasing cognitive complexity. The colored texts in ``Scenario Description" correspond to the same colors in ``Rule Basis", as well as the colored boxes in the visual context, indicating the rule-to-scene alignment.}
    \label{fig: qa_example}
\end{figure*}
\section{DriveCombo}

\subsection{Dataset Overview}

The DriveCombo dataset is designed to evaluate models’ capability in understanding traffic rules and performing visual–language reasoning in driving scenarios. It contains about 70K multiple-choice questions (MCQs) across driving topics such as traffic signals, vehicle operations, emergency yielding, and headlight use. The whold benchmark spans five cognitive levels:

\begin{itemize}
    \item \textbf{L1:} Evaluates understanding of basic atomic traffic rules.
    \item \textbf{L2:} Tests the ability to integrate multiple non-conflicting static rules (for example, road signs or speed limits).
    \item \textbf{L3:} Assesses reasoning under dynamic interactions involving other road participants.
    \item \textbf{L4:} Examines mixed scenarios that combine both static and dynamic rules.
    \item \textbf{L5:} Assesses whether a model prioritizes higher-level obligations when rules conflict.
\end{itemize}

\begin{figure*}[ht!]
    \centering
    \vspace{-0.4cm}
      \includegraphics[width=0.96\linewidth]{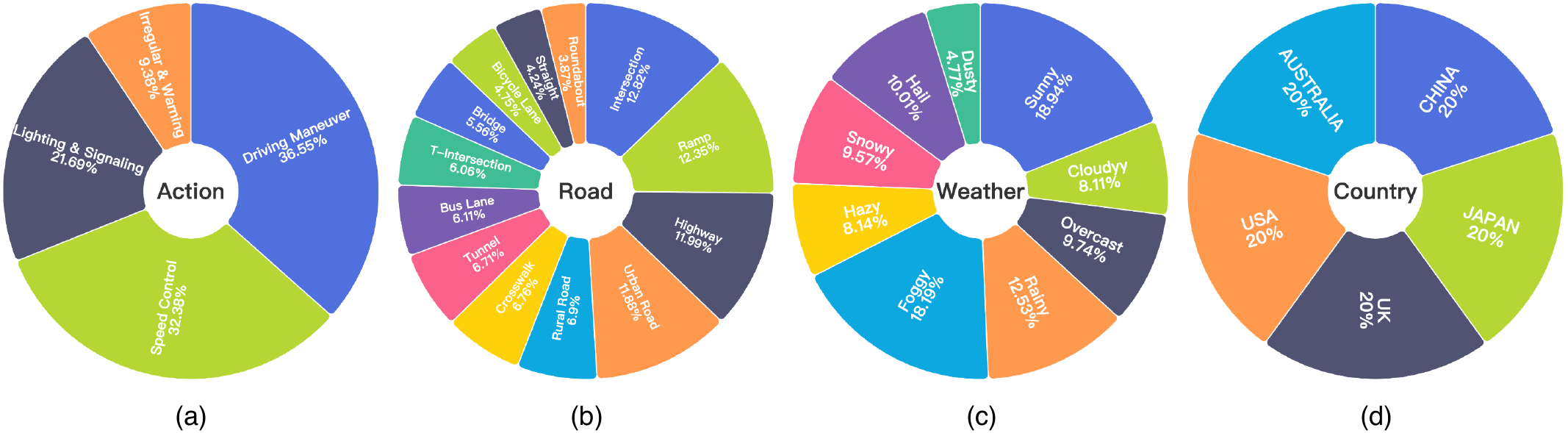}
    \caption{ \textbf{Data Distribution} across \textbf{(a)} action types, \textbf{(b)} road types, \textbf{(c)} weather conditions, and \textbf{(d)} country sources in DriveCombo. 
    }
    \label{fig: distribution} 
\end{figure*}

Figure~\ref{fig: qa_example} illustrates the task examples for five levels above. 
Figure~\ref{fig: distribution} summarizes the key statistics of the DriveCombo dataset. The distribution of driving action types is balanced, with driving maneuvers and speed control being the most common. The dataset covers more than ten diverse road types and includes various eight weather conditions, ensuring robustness under different visibility conditions. The dataset covers five countries, demonstrating broad regional and regulatory coverage.

\subsection{Dataset Construction}

To build a hierarchical rule base for five-level question design and scenario generation, we hierarchically model traffic rules to generate textual and visual scenarios, which are then assembled into multi-level MCQs as described below.

\paragraph{Atomic Traffic Rules Extraction.} Firstly, we collect and parse official driving manuals and traffic codes from 5 countries.  Through LLM-assisted syntactic analysis and human verification, original compound rules are decomposed into atomic rules, forming a set \( R = \{r_i\}_{i=1}^{|R|} \) for each country. Each atomic rule \( r_i \) represents a well-defined “situation–action” mapping (e.g., “When warning lights flash, the driver must yield to emergency vehicles”). This atomic decomposition ensures semantic clarity and supports fine-grained reasoning.

\paragraph{Scene Generation.} Next, our Rule2Scene Agent (see Section~\ref{sec:rule2sceneagent}), a human–AI collaborative pipeline, generate scenario texts that encode each rule’s semantics and constraints, and these are subsequently simulated to create visual data supporting both question narratives and evaluation inputs.  At each stage, another LLM performs a quality check and scores the output. If the score falls below a threshold, human experts refine the result to ensure realism and strict compliance with the intended rule semantics.

\paragraph{Construction of Multi-Level Questions.} Each question comprising a scenario description, stem, and four options for evaluation (see Appendix Section~1). The correct answer is concluded by priority principle~\cite{china_2021_road_traffic_safety_law, china_2017_road_traffic_safety_law_regulations}, while the other three distractors are LLM-generated and human-screened for semantic plausibility and controlled deviation.

\section{Rule2Scene Agent}

\begin{figure*}[ht!]
    \centering
    \vspace{-0.5cm}
      \includegraphics[width=1.0\linewidth]{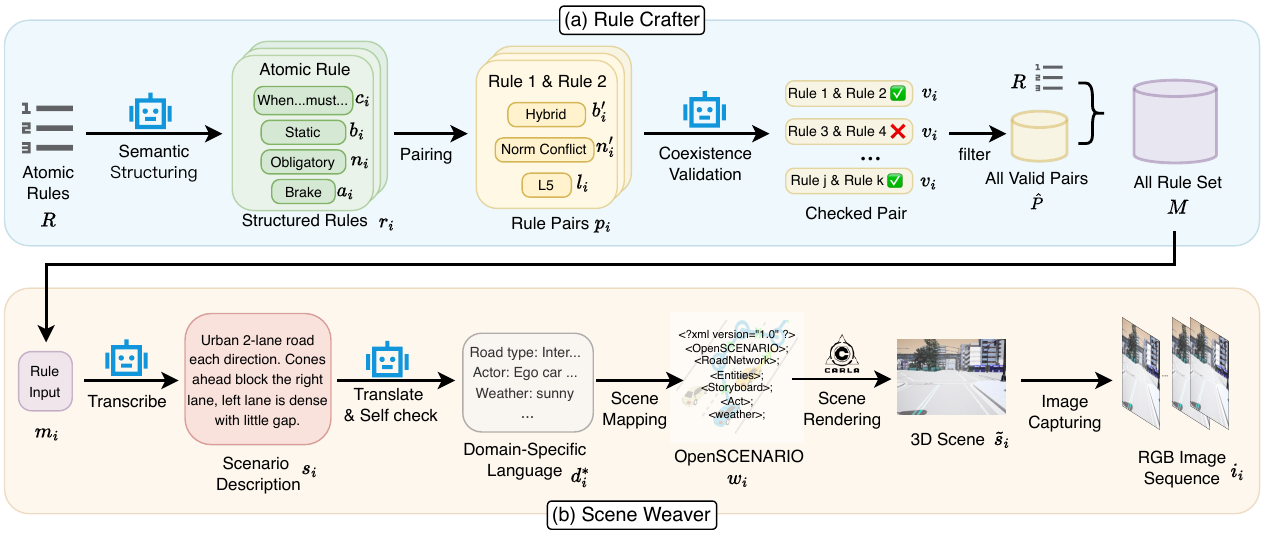}
    \caption{\textbf{Rule2Scene Agent}. 
    The agent consists of two modules: \textbf{(a) Rule Crafter} performs semantic structuring of atomic rules \( r_i \in R\), generates candidate rule pairs $p_i$, verifies spatiotemporal coexistence to construct a hierarchical rule set $M$; \textbf{(b) Scene Weaver} converts the hierarchical rules $m_i \in M$ into textual scene descriptions $s_i$, generates structured semantic representations $w_i$, maps them to the CARLA simulator~\cite{dosovitskiy2017carla}, renders and captures images $i_i$, and finally generates high-fidelity driving scenarios for model evaluation.}
    \label{fig: method_Dataset_Construction} 
\end{figure*}

To generate rule-related question texts and high-fidelity visual inputs, we develop the Rule2Scene Agent, whose overall framework is illustrated in Figure~\ref{fig: method_Dataset_Construction}. The system comprises two modules: the Rule Crafter and the Scene Weaver.

\subsection{Rule Crafter}
To generate high-quality questions, we first employ the Rule Crafter to model atomic rules into a hierarchical rule set.

\paragraph{LLM-Based Semantic Structuring.}
Each rule is semantically structured as \( r_i = (c_i, b_i, a_i, n_i) \),  
where \( c_i \) is the rule content, \( b_i \in B \) the perceptual type, \( a_i \in A \) the action type, and \( n_i \in N \) the norm type.  
The three state spaces are defined as  
\( B = \{\text{static}, \text{dynamic}\} \),  
\( A = \{\text{overtake}, \text{accelerate}, \text{u-turn}, …\} \), and  
\( N = \{\text{permissive}, \text{obligatory}, \text{forbidden}\} \).  

\paragraph{Candidate Rule Pairs Generation.}
To capture semantic and normative relationships among rules, action type consistent pairs are formed as  
\( P = \{p_i = \{r_j, r_k\} \mid a_j = a_k,\, r_j, r_k \in R\}_{i=1}^{|P|} \).  
Each pair \( p_i \) is assigned three derived labels: the perceptual combination type \( b_i' \), the normative relation type \( n_i' \), and the hierarchical level \( l_i \):  
\[
b_i' =
\begin{cases}
\text{Double Static}, & b_j=b_k=\text{static},\\
\text{Double Dynamic}, & b_j=b_k=\text{dynamic},\\
\text{Hybrid}, & \text{otherwise.}
\end{cases}
\]
\[
n_i' =
\begin{cases}
\text{Norm Conflict}, & \{n_j,n_k\}=\{\text{obligatory},\text{forbidden}\},\\
\text{Norm Harmony}, & \text{otherwise.}
\end{cases}
\]
\[
l_i =
\begin{cases}
2, & b_i'=\text{Double Static}, n_i'=\text{Norm Harmony},\\
3, & b_i'=\text{Double Dynamic}, n_i'=\text{Norm Harmony},\\
4, & b_i'=\text{Hybrid}, n_i'=\text{Norm Harmony},\\
5, & n_i'=\text{Norm Conflict}.
\end{cases}
\]
All atomic rules \( r_i \in R \) are labeled \( l_i = 1 \), forming the complete hierarchical structure (L1–L5). Note that rules related to speed limits are handled with additional logic (see Section 1 of the Appendix).

\paragraph{Spatio-Temporal Coexistence Validation.}
Each candidate rule pair is first verified by the LLM to assess whether the two rules can coexist within the same physical and temporal context.  During this process, the model extracts contextual elements from each rule, such as road type, agent state, and environmental attributes, and filters out pairs that are semantically and physically incompatible. Each candidate pair \( p_i = \{r_j, r_k\} \) is validated through the function \( v_i = f_{\text{LLM}}(r_j, r_k) \), where \( v_i \in \{0,1\} \) denotes coexistence feasibility, with \( v_i = 1 \) indicating that the rules can coexist and \( v_i = 0 \) otherwise.

Finally, All valid rule pairs are retained in \( \hat{P} = \{p_i \mid v_i = 1\} \),  
and the final rules set is obtained as \( M = R \cup \hat{P} \),  
covering all levels from L1 to L5. 

\label{sec:rule2sceneagent}

\subsection{Scene Weaver}
After generating the hierarchical rules, we employ the Scene Weaver module to convert them into textual scene descriptions, further producing executable simulation programs and producing visualized data to obtain high-fidelity driving scenarios for model evaluation.

\paragraph{Textual Scene Description Generation.} To transform abstract rules into perceptible driving contexts, we first employ LLM to transcribe each input \( m_i \in M \) into a textual scenario description \( s_i \). Based on the number of rules and the semantic interrelations among them, the model generates integrated contextual scenarios that reflect the single rule or multiple rules and their constraints simultaneously.

\paragraph{Structured Scene Generation.}
We then use LLM to translate the scenario description \( s_i \) into a structured semantic representation 
\( d_i = \{E_i, L_i, W_i\} \), where \( E_i \) denotes entities (e.g., vehicles, pedestrians, lights), 
\( L_i \) represents spatial or interaction relations (e.g., ``ahead,'' ``left of''), 
and \( W_i \) captures environmental conditions (e.g., weather, time, road type). This structure follows a Domain-Specific Language (DSL) based on the traffic simulation schema proposed in Target~\cite{deng2025target}.  A multi-stage pipeline, consisting of the stages \textit{Generate \(\rightarrow\) SelfCheck \(\rightarrow\) Align}, ensures that the resulting structure \( d_i^\ast \) preserves syntactic and semantic coherence.

\paragraph{Scene Mapping in Simulator.}
The structured semantics \( d_i^{\ast} \) are mapped into CARLA~\cite{dosovitskiy2017carla}’s 3D coordinate space to create OpenSCENARIO~\cite{openScenario} file \( \omega_i \), including entity positions, traffic structures, weather conditions, and entity trajectories. All of this information was obtained through interactions with CARLA, and the detailed implementation process is described in Appendix Section 1.

\paragraph{Scene Rendering and Capturing.}
To obtain high-quality visual inputs, each OpenSCENARIO file $\omega_i$ was imported into the CARLA simulator and rendered into a 3D scene $\tilde{s}i$. Subsequently, a camera was positioned in front of the ego vehicle to capture a sequence of $K$ RGB image frames, denoted as $i_i = { i_i^{(1)}, i_i^{(2)}, \dots, i_i^{(K)} }$. At this stage, the visual scene set \( I = \{ i_i \}_{i=1}^{|I|} \) and textual scene set \( S = \{ s_i \}_{i=1}^{|S|} \) are fully prepared for subsequent evaluation.

\section{Experiments}
\label{experiments}

\subsection{Experiment Setup}

\paragraph{Model Selection.} We systematically evaluated 14 MLLMs, covering a diverse range of architectures and parameter scales. The proprietary models included the GPT-5 series (nano, mini, pro)~\cite{openai2025gpt5}, Gemini-2.5-Flash~\cite{team2023gemini}, Gemini-2.5-Pro~\cite{team2023gemini}, and Claude-Sonnet-4.5~\cite{anthropic2024Claude-Sonnet-4.5}. In addition, several open-source models were evaluated, including the Gemma-3 series~\cite{team2025gemma3}, Llama-3.2~\cite{zhang2023Llama}, Qwen-3-VL series~\cite{yang2025qwen3}, and GLM-4.5V~\cite{vteam2025glm45v}. 

\paragraph{Data Split.}
We used all atomic traffic rules and 80\% of the driving test samples to construct the training set, reserving the remaining 20\% for evaluation. From the same test set, we derived the DriveCombo-Text variant by replacing visual frames with textual scene descriptions, enabling an isolated assessment of text-based reasoning without visual input.

\paragraph{Metrics.}
We use accuracy as the primary evaluation metric to measure model’s ability to answer driving-related questions correctly. All questions were multiple-choice (MCQs) with a single correct answer. 

\paragraph{Implementation Details.} Rules from different countries are constructed and evaluated independently without cross-country mixing. For each MCQ, we capture four RGB frames as the visual inputs (e.g., $K=4$). The detailed action types in the action space $A$ and the priority principles are provided in Sec. 1 of the Appendix. All evaluations are conducted in a zero-shot setting. To ensure robustness and statistical significance, each model was independently run 3 times on the same test set for reproducibility. We also invited 30 human drivers to manually complete a set of 100 randomly selected questions from each difficulty level in the test set. Their accuracy exceeded 98\%, confirming the reliability of our benchmark.

\paragraph{Reasoning Augmentation.} To evaluate the effectiveness of various reasoning-enhancement methods within DriveCombo, we test three enhancement strategies: Chain-of-Thought (CoT)~\cite{wei2022cot} for explicit reasoning, Retrieval-Augmented Generation (RAG)~\cite{lewis2020rag} for external knowledge grounding, and Supervised Fine-Tuning (SFT) for supervised adaptation. The specific implementation details of SFT, CoT and RAG are provided in Sec. 2 of the Appendix.

\subsection{Main Results}
\paragraph{Performance of MLLMs on DriveCombo. }
Table~\ref{tab:DriveCTR_result} presents the performance of MLLMs on the DriveCombo. Proprietary models consistently outperform open-source models, yet both groups exhibit similar performance trends across cognitive levels. All models achieve high accuracy on L1 atomic rule tasks but show a steady decline in L2–L4 multi-rule reasoning. The most challenging L5 priority conflict arbitration tasks cause accuracy to drop sharply to 41\%–44\%, well below human performance. These findings highlight the current limitations of MLLMs in handling hierarchical rule conflicts and validate the effectiveness of our five-level cognitive ladder in revealing reasoning depth. The results underscore the necessity of structured, hierarchical datasets to advance reasoning for real-world driving.

\begin{table}[t!]
\caption{\textbf{Performance of MLLMs on DriveCombo.} 
\colorbox[HTML]{ADD88D}{Green} and \colorbox[HTML]{E3F2D9}{light green} mark the best and second-best open-source models, while \colorbox[HTML]{B5C6EA}{blue} and \colorbox[HTML]{D9E1F4}{light blue} indicate the best and second-best proprietary models.
}
\label{tab:DriveCTR_result}
\centering
\footnotesize
\setlength{\tabcolsep}{5pt}
\begin{tabular}{lc|ccccc}
\toprule
\textbf{Model}             & \textbf{Size} & \textbf{L1}                   & \textbf{L2}                   & \textbf{L3}                   & \textbf{L4}                   & \textbf{L5}                   \\
\midrule
\multicolumn{7}{c}{\textit{Open-source Models}}                                                                                                                                                            \\
Gemma 3           & 4B   & 58.79                         & 47.80                         & 45.98                         & 44.55                         & 26.66                         \\
Gemma 3           & 12B  & 64.32                         & 58.69                         & 54.90                         & 52.32                         & 29.54                         \\
Gemma 3           & 27B  & 73.94                         & 67.39                         & 65.55                         & 63.10                         & 37.42                         \\
Llama 3.2         & 11B  & 52.46                         & 49.23                         & 45.75                         & 41.94                         & 25.74                         \\
Qwen3-VL          & 2B   & 61.72                         & 59.29                         & 57.39                         & 53.15                         & 25.10                         \\
Qwen3-VL          & 8B   & 72.46                         & 66.63                         & 63.78                         & 55.39                         & 35.03                         \\
Qwen3-VL          & 32B  & \cellcolor[HTML]{E3F2D9}78.54 & \cellcolor[HTML]{E3F2D9}76.09 & \cellcolor[HTML]{E3F2D9}68.84 & \cellcolor[HTML]{E3F2D9}65.42 & \cellcolor[HTML]{E3F2D9}39.86 \\
GLM-4.5V          & 106B & \cellcolor[HTML]{ADD88D}80.44 & \cellcolor[HTML]{ADD88D}78.49 & \cellcolor[HTML]{ADD88D}69.54 & \cellcolor[HTML]{ADD88D}68.22 & \cellcolor[HTML]{ADD88D}41.86 \\
\midrule
\multicolumn{7}{c}{\textit{Proprietary Models}}                                                                                                                                                            \\
Gemini 2.5 Flash  & -             & 79.31                         & 72.07                         & 67.59                         & 64.03                         & 39.33                         \\
Gemini 2.5 Pro    & -             & \cellcolor[HTML]{D9E1F4}85.71 & 77.19                         & 70.32                         & 68.03                         & 43.06                         \\
Claude Sonnet 4.5 & -             & 83.80                         & \cellcolor[HTML]{D9E1F4}82.16 & \cellcolor[HTML]{D9E1F4}70.96 & \cellcolor[HTML]{D9E1F4}69.62 & \cellcolor[HTML]{D9E1F4}43.99
\\       
GPT-5 nano        & -             & 74.04                         & 69.76                         & 67.09                         & 63.43                         & 33.24                         \\
GPT-5 mini        & -             & 80.62                         & 72.59                         & 69.61                         & 62.27                         & 38.10                         \\
GPT-5 pro         & -             & \cellcolor[HTML]{B5C6EA}86.91 & \cellcolor[HTML]{B5C6EA}83.66 & \cellcolor[HTML]{B5C6EA}72.06 & \cellcolor[HTML]{B5C6EA}69.82 & \cellcolor[HTML]{B5C6EA}44.19 \\
\bottomrule
\end{tabular}
\end{table}

\begin{table}[t!]
\caption{\textbf{Performance of MLLMs on DriveCombo-Text variant (only Text Input).} 
}
\label{tab:DriveCTR_Text_result}
\footnotesize
\setlength{\tabcolsep}{5pt}
\begin{tabular}{lc|ccccc}
\toprule
\textbf{Model}    & \textbf{Size} & \textbf{L1}                   & \textbf{L2}                   & \textbf{L3}                   & \textbf{L4}                   & \textbf{L5}                   \\
\midrule
\multicolumn{7}{c}{\textit{Open-source Models}}                                                                                                                                                   \\
Gemma 3           & 4B            & 62.61                         & 51.75                         & 50.93                         & 48.16                         & 29.82                         \\
Gemma 3           & 12B           & 68.28                         & 62.19                         & 59.39                         & 56.56                         & 33.08                         \\
Gemma 3           & 27B           & 78.30                         & 71.20                         & 69.51                         & 67.55                         & 41.15                         \\
Llama 3.2         & 11B           & 58.15                         & 53.89                         & 51.36                         & 45.59                         & 30.22                         \\
Qwen3-VL          & 2B            & 66.67                         & 63.08                         & 61.40                         & 57.47                         & 29.07                         \\
Qwen3-VL          & 8B            & 77.61                         & 70.13                         & 67.92                         & 59.54                         & 38.78                         \\
Qwen3-VL          & 32B           & \cellcolor[HTML]{E3F2D9}83.28 & \cellcolor[HTML]{E3F2D9}79.58 & \cellcolor[HTML]{E3F2D9}72.90 & \cellcolor[HTML]{E3F2D9}70.04 & \cellcolor[HTML]{E3F2D9}43.33 \\
GLM-4.5V          & 106B          & \cellcolor[HTML]{ADD88D}85.77 & \cellcolor[HTML]{ADD88D}83.04 & \cellcolor[HTML]{ADD88D}75.07 & \cellcolor[HTML]{ADD88D}71.94 & \cellcolor[HTML]{ADD88D}45.49 \\
\midrule
\multicolumn{7}{c}{\textit{Proprietary Models}}                                                                                                                                                   \\
Gemini 2.5 Flash  & -             & 82.84                         & 75.42                         & 72.04                         & 69.33                         & 42.77                         \\
Gemini 2.5 Pro    & -             & \cellcolor[HTML]{D9E1F4}88.92 & 80.88                         & 75.41                         & 73.76                         & 46.60                         \\
Claude Sonnet 4.5 & -             & 88.63                         & \cellcolor[HTML]{D9E1F4}86.38 & \cellcolor[HTML]{B5C6EA}77.90 & \cellcolor[HTML]{D9E1F4}74.38 & \cellcolor[HTML]{D9E1F4}46.82
\\
GPT-5 nano        & -             & 77.85                         & 73.50                         & 72.42                         & 66.82                         & 36.33                         \\
GPT-5 mini        & -             & 85.04                         & 77.17                         & 73.44                         & 66.95                         & 41.47                         \\
GPT-5 pro         & -             & \cellcolor[HTML]{B5C6EA}89.23 & \cellcolor[HTML]{B5C6EA}86.96 & \cellcolor[HTML]{D9E1F4}77.53 & \cellcolor[HTML]{B5C6EA}75.78 & \cellcolor[HTML]{B5C6EA}47.42 \\       
\bottomrule
\end{tabular}
\end{table}

\paragraph{Performance of MLLMs on DriveCombo-Text variant.}
Table~\ref{tab:DriveCTR_Text_result} presents model results on the DriveCombo-Text variant. Compared with the visual version, all models show a modest improvement in accuracy, suggesting that removing visual perception pressure allows for clearer textual reasoning. This improvement indicates that the visual understanding stage still suffers from semantic loss and misalignment. Nonetheless, even under text-only variant, the best-performing model achieves only 47.42\% accuracy on L5 conflict arbitration task, reaffirming that MLLMs remain limited in high-level priority-based reasoning.

\paragraph{Reasoning Augmentation Performance.}
The above analysis reveal substantial room for improvement in complsitional traffic rule reasoning. To explore whether conventional optimization strategies can mitigate this gap, we conducted experiments on five open-source models, Gemma-3 series, Qwen3-VL series, and Llama-3.2, which initially performed poorly and span different parameter scales. We compared training-free approaches (CoT and RAG) with training-dependent strategies (SFT). As shown in Table~\ref{tab:augment_result_CoT_RAG_SFT}, SFT outperformed CoT and RAG, significantly boosting reasoning accuracy in multi-rule composition tasks. For example, Llama-3.2 achieved a 29.7\% accuracy gain after SFT on DriveCombo, demonstrating that our dataset effectively injects transferable traffic rule knowledge into MLLMs. However, even the best fine-tuned model, Qwen3-VL-8B, reached only 60.2\% accuracy on L5 tasks, which still far below human levels. This highlights limitations of conventional optimization in complex rule reasoning.

\begin{table}[t!]
\caption{
\textbf{Evaluating DriveCombo as a knowledge injector for MLLMs} with three enhancement strategies (CoT, RAG, and SFT).
}
\label{tab:augment_result_CoT_RAG_SFT}
\centering
\footnotesize
\setlength{\tabcolsep}{2pt}
\begin{tabular}{lc|ccc|ccccc|c}
\toprule
\textbf{Model}                     & \textbf{Size}                 & \textbf{CoT}              & \textbf{RAG}              & \textbf{SFT}         & \textbf{L1} & \textbf{L2} & \textbf{L3} & \textbf{L4} & \textbf{L5} & \textbf{Avg.} \\
\midrule
\multirow{4}{*}{Gemma 3}    & \multirow{4}{*}{4B}  &                           &                           &                           & 58.8      & 47.8      & 46.0      & 44.5      & 26.7      & 44.8           \\
                                    &                               & \checkmark &                           &                           & 60.1      & 52.4      & 48.3      & 47.2      & 29.2      & +2.70           \\
                                    &                               & \checkmark & \checkmark &                           & 63.4      & 57.6      & 54.2      & 51.2      & 33.4      & +7.31           \\
                                    &                               & \checkmark & \checkmark & \checkmark & 86.3      & 81.9      & 76.9      & 74.2      & 51.3      & +29.37           \\
\midrule
\multirow{4}{*}{Gemma 3}    & \multirow{4}{*}{12B} &                           &                           &                           & 64.3      & 58.7      & 54.9      & 52.3      & 29.5      & 51.9           \\
                                    &                               & \checkmark &                           &                           & 65.0      & 60.4      & 56.2      & 53.2      & 30.3      & +1.06           \\
                                    &                               & \checkmark & \checkmark &                           & 67.3      & 62.2      & 59.7      & 54.5      & 34.7      & +3.72           \\
                                    &                               & \checkmark & \checkmark & \checkmark & 89.8      & 85.9      & 78.0      & 76.4      & 58.2      & +25.73           \\
\midrule
\multirow{4}{*}{Qwen3-VL}  & \multirow{4}{*}{2B}  &                           &                           &                           & 61.7      & 59.3      & 57.4      & 53.1      & 25.1      & 51.3           \\
                                    &                               & \checkmark &                           &                           & 63.2      & 61.7      & 58.2      & 54.1      & 28.3      & +1.77          \\
                                    &                               & \checkmark & \checkmark &                           & 66.4      & 64.5      & 59.8      & 55.2      & 33.5      & +4.56           \\
                                    &                               & \checkmark & \checkmark & \checkmark & 86.8      & 83.0      & 78.0      & 74.6      & 54.7      & +24.09           \\
\midrule
\multirow{4}{*}{Qwen3-VL}  & \multirow{4}{*}{8B}  &                           &                           &                           & 72.5      & 66.6      & 63.8      & 55.4      & 35.0      & 58.7           \\
                                    &                               & \checkmark &                           &                           & 74.0      & 67.2      & 65.3      & 57.6      & 36.8      & +1.52           \\
                                    &                               & \checkmark & \checkmark &                           & 75.8      & 69.1      & 68.4      & 59.7      & 40.9      & +4.13           \\
                                    &                               & \checkmark & \checkmark & \checkmark & 90.3      & 89.8      & 83.7      & 78.7      & 60.2      & +21.89           \\
\midrule
\multirow{4}{*}{Llama 3.2} & \multirow{4}{*}{11B} &                           &                           &                           & 52.5      & 49.2      & 45.7      & 41.9      & 25.7      & 43.0           \\
                                    &                               & \checkmark &                           &                           & 55.2      & 52.9      & 47.3      & 43.8      & 29.0      & +2.62           \\
                                    &                               & \checkmark & \checkmark &                           & 58.0      & 54.5      & 50.8      & 45.7      & 33.2      & +5.42           \\
                                    &                               & \checkmark & \checkmark & \checkmark & 85.6      & 80.5      & 75.2      & 72.3      & 50.1      & +29.70         
\\
\bottomrule
\end{tabular}
\end{table}

\paragraph{Benefits of DriveCombo for Downstream E2E Planning task.}
Table~\ref{tab:planning_result} reports the results of several MLLMs on downstream end-to-end trajectory planning on the nuScenes dataset. We compared model performance under DriveQA fine-tuning, and DriveCombo fine-tuning conditions. Models fine-tuned on DriveCombo achieved substantially lower L2 trajectory errors, indicating a closer alignment with expert trajectories. These results confirm the practical value of our dataset in enhancing the planning accuracy and safety of MLLMs for real-world driving applications.

\begin{table}[t!]
\caption{\textbf{End-to-End Trajectory Planning Results on nuScenes validation set.} We compute the L2 error at different prediction horizons (1s, 2s, and 3s). A lower L2 error indicates that our DriveCombo dataset transfers from simulation to real-world planning tasks more effectively than DriveQA~\cite{wei2025driveqa}.}
\label{tab:planning_result}
\centering
\footnotesize
\setlength{\tabcolsep}{4.5pt}
\begin{tabular}{l|c|cccc}
\toprule
                                                &                                          & \multicolumn{4}{c}{\textbf{L2($m$) $\downarrow$}}                                                                                       \\
\multirow{-2}{*}{\textbf{Model}}                & \multirow{-2}{*}{\textbf{SFT Data}} & 1$s$                  & 2$s$                  & 3$s$                  & Avg.                \\
\midrule
                                                & -                               & 1.66                         & 3.54                         & 4.54                         & 3.24                         \\
                                                & DriveQA                         & 1.30                         & 3.46                         & 3.98                         & 2.91                         \\
\multirow{-3}{*}{LlaVa-1.6-Mistral-7B~\cite{li2024llava}} & DriveCombo                        & \cellcolor[HTML]{E3F2D9}1.27 & 3.29                         & 3.92                         & \cellcolor[HTML]{E3F2D9}2.68 \\
\midrule
                                                & -                               & 1.66                         & 3.36                         & 4.15                         & 3.06                         \\
                                                & DriveQA                              & 1.30                         & \cellcolor[HTML]{E3F2D9}3.08 & \cellcolor[HTML]{E3F2D9}3.73 & 2.71                         \\
\multirow{-3}{*}{InternVL-2.5-8B~\cite{gao2024InternVL}}      & DriveCombo                        & \cellcolor[HTML]{ADD88D}1.26 & \cellcolor[HTML]{ADD88D}3.03 & \cellcolor[HTML]{ADD88D}3.58 & \cellcolor[HTML]{ADD88D}2.53
\\
\bottomrule
\end{tabular}
\end{table}

\begin{figure}[t!]
    \centering
    \vspace{0.15cm}
      \includegraphics[width=1.0\linewidth]{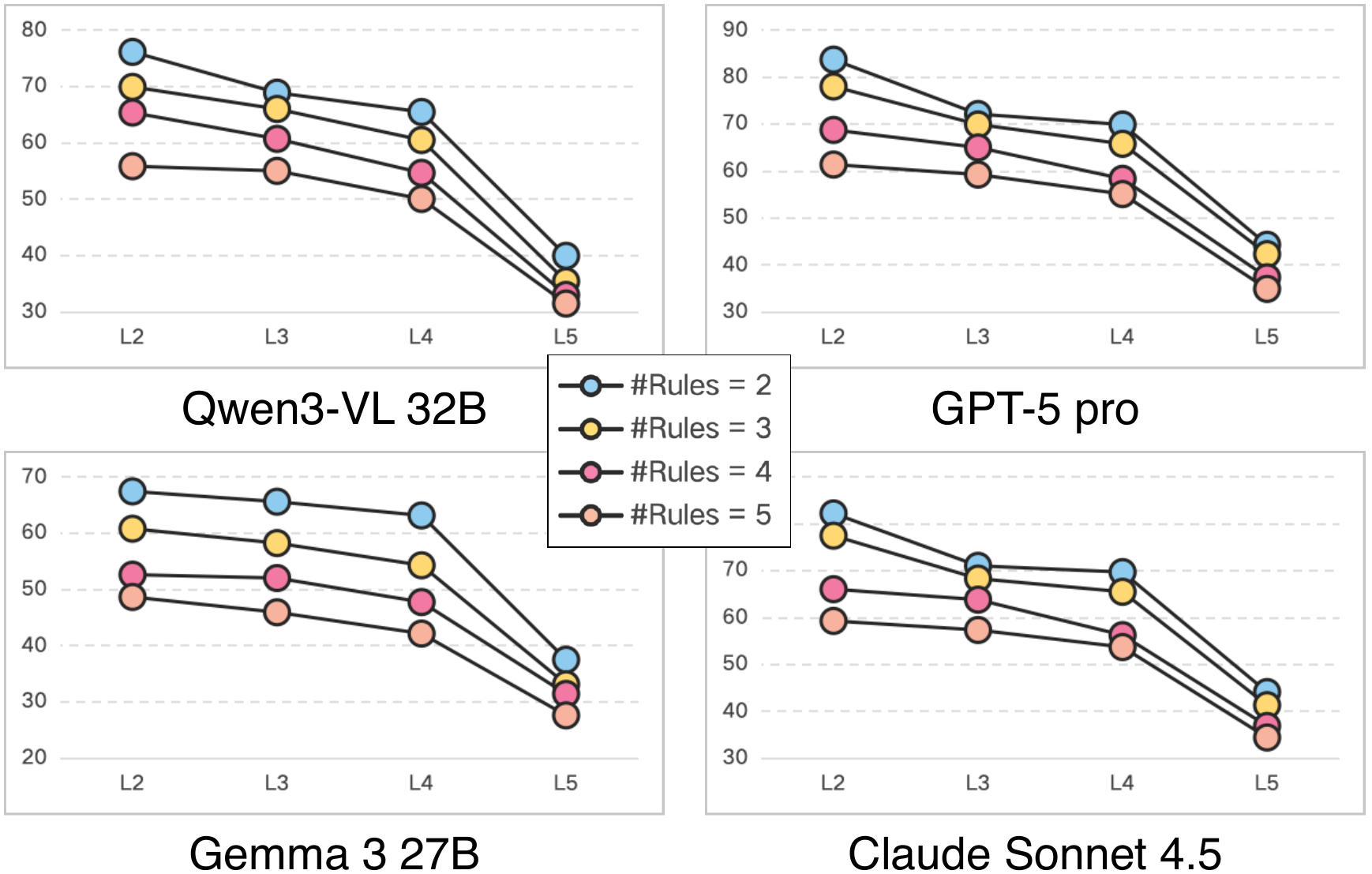}
    \caption{ \textbf{Performance of MLLMs on Multi-Rule Compositional Reasoning}. ``\#Rules" means the number of traffic rules within each scenario. The results are obtained in a zero-shot setting. Since L1 is single-rule setting, we only present L2-L5 here.
    }
    \label{fig: comb_3_4_5} 
    \vspace{0.1cm}
\end{figure}

\paragraph{Complexity Analysis of Multi-Rule Compositional Reasoning.}
To further assess the model’s reasoning ability under higher-complexity conditions, we additionally conducted in-depth experiments on scenarios involving combinations of more than two rules, systematically examining model performance under higher-order rule interactions. Figure~\ref{fig: comb_3_4_5} illustrates the performance across different numbers of rule within the L2–L5 levels. The results indicate that as the number of rules involved in each scenario increases, the accuracy of all models declines significantly, which is evident at higher cognitive levels (especially L4 and L5). Specifically, even large-scale models such as GPT-5 pro and Qwen3-VL 32B show strong performance under two-rule settings, but their accuracy drops by over 20 percentage points when the rules increase to four or five. This trend indicates that more rule combinations significantly raise cognitive load, making it hard for models to sustain consistent reasoning in high-dimensional traffic semantics. This further underscores the importance of DriveCombo’s hierarchical cognitive design, which systematically quantifies and reveals the inherent limitations of MLLMs in multi-rule compositional reasoning. More quantitative results are provided in Sec. 3 of the Appendix.
\section{Conclusion}
This paper introduces DriveCombo, a benchmark for compositional traffic rule reasoning. DriveCombo defines a five-level cognitive hierarchy, progressing from single-rule understanding to multi-rule conflict resolution, and a Rule2Scene Agent that converts textual rules into dynamic driving scenes, enabling cross-modal evaluation. Experiments of 14 MLLMs reveal substantial performance drops in conflict resolution, highlighting gaps in complex reasoning. Furthermore, fine-tuning on DriveCombo enhances end-to-end planning, demonstrating its potential to foster compliant driving agents.
Although current scene generation is constrained by CARLA’s asset library~\cite{dosovitskiy2017carla}, we plan to integrate generative models~\cite{song2025insertany} to expand its evaluation scope. Overall, DriveCombo advances the development of reliable and compliant multimodal driving agents.


\section*{Acknowledgments}
This work is partially supported by the National Natural Science Foundation of China (NSFC) under grant No. 62403389, the Provincial Natural Science Foundation of Zhejiang under grant No. QKWL25F0301, and the Zhejiang Key Laboratory of Low-Carbon Intelligent Synthetic Biology (2024ZY01025).
{    
\small
    \bibliographystyle{ieeenat_fullname}
    \bibliography{main}

@String(CVPR= {IEEE Conf. Comput. Vis. Pattern Recog.})

@String(ICCV= {Int. Conf. Comput. Vis.})

@String(ECCV= {Eur. Conf. Comput. Vis.})

@String(ICLR = {Int. Conf. Learn. Represent.})

@String(AAAI = {AAAI})

@String(CVPR  = {CVPR})

@String(ICCV  = {ICCV})

@String(ECCV  = {ECCV})

@String(ICLR  = {ICLR})

@article{jia2024bench2drive,
  title={Bench2drive: Towards multi-ability benchmarking of closed-loop end-to-end autonomous driving},
  author={Jia, Xiaosong and Yang, Zhenjie and Li, Qifeng and Zhang, Zhiyuan and Yan, Junchi},
  journal={Advances in Neural Information Processing Systems},
  volume={37},
  pages={819--844},
  year={2024}
}

@inproceedings{hu2023uniad,
  title={Planning-oriented autonomous driving},
  author={Hu, Yihan and Yang, Jiazhi and Chen, Li and Li, Keyu and Sima, Chonghao and Zhu, Xizhou and Chai, Siqi and Du, Senyao and Lin, Tianwei and Wang, Wenhai and others},
  booktitle={Proceedings of the IEEE/CVF Conference on Computer Vision and Pattern Recognition},
  pages={17853--17862},
  year={2023}
}

@article{Jiang2023VADVS,
  title={VAD: Vectorized Scene Representation for Efficient Autonomous Driving},
  author={Bo Jiang and Shaoyu Chen and Qing Xu and Bencheng Liao and Jiajie Chen and Helong Zhou and Qian Zhang and Wenyu Liu and Chang Huang and Xinggang Wang},
  journal={2023 IEEE/CVF International Conference on Computer Vision (ICCV)},
  year={2023},
  pages={8306-8316}
}

@article{zhou2024embodied,
  title={Embodied understanding of driving scenarios},
  author={Zhou, Yunsong and Huang, Linyan and Bu, Qingwen and Zeng, Jia and Li, Tianyu and Qiu, Hang and Zhu, Hongzi and Guo, Minyi and Qiao, Yu and Li, Hongyang},
  journal={ECCV},
  year={2024}
}

@article{wen2023dilu,
  title={Dilu: A knowledge-driven approach to autonomous driving with large language models},
  author={Wen, Licheng and Fu, Daocheng and Li, Xin and Cai, Xinyu and Ma, Tao and Cai, Pinlong and Dou, Min and Shi, Botian and He, Liang and Qiao, Yu},
  journal={arXiv:2309.16292},
  year={2023}
}

@article{li2024llava,
  title={Llava-next-interleave: Tackling multi-image, video, and 3d in large multimodal models},
  author={Li, Feng and Zhang, Renrui and Zhang, Hao and Zhang, Yuanhan and Li, Bo and Li, Wei and Ma, Zejun and Li, Chunyuan},
  journal={arXiv preprint arXiv:2407.07895},
  year={2024}
}

@article{gao2024InternVL,
  title={Mini-InternVL: a flexible-transfer pocket multi-modal model with 5\% parameters and 90\% performance},
  author={Gao, Zhangwei and Chen, Zhe and Cui, Erfei and Ren, Yiming and Wang, Weiyun and Zhu, Jinguo and Tian, Hao and Ye, Shenglong and He, Junjun and Zhu, Xizhou and others},
  journal={Visual Intelligence},
  year={2024}
}

@article{sha2023languagempc,
  title={Languagempc: Large language models as decision makers for autonomous driving},
  author={Sha, Hao and Mu, Yao and Jiang, Yuxuan and Chen, Li and Xu, Chenfeng and Luo, Ping and Li, Shengbo Eben and Tomizuka, Masayoshi and Zhan, Wei and Ding, Mingyu},
  journal={arXiv:2310.03026},
  year={2023}
}

@article{mao2023gpt,
  title={GPT-Driver: Learning to Drive with GPT},
  author={Mao, Jiageng and Qian, Yuxi and Zhao, Hang and Wang, Yue},
  journal={arXiv:2310.01415},
  year={2023}
}

@inproceedings{fu2024drive,
  title={Drive like a human: Rethinking autonomous driving with large language models},
  author={Fu, Daocheng and Li, Xin and Wen, Licheng and Dou, Min and Cai, Pinlong and Shi, Botian and Qiao, Yu},
  booktitle={WACV},
  year={2024}
}

@article{cui2024receive,
  title={Receive, reason, and react: Drive as you say, with large language models in autonomous vehicles},
  author={Cui, Can and Ma, Yunsheng and Cao, Xu and Ye, Wenqian and Wang, Ziran},
  journal={IEEE ITS Magazine},
  year={2024},
}

@inproceedings{cui2024drive,
  title={Drive as you speak: Enabling human-like interaction with large language models in autonomous vehicles},
  author={Cui, Can and Ma, Yunsheng and Cao, Xu and Ye, Wenqian and Wang, Ziran},
  booktitle={WACV},
  year={2024}
}

@inproceedings{chen2024driving,
  title={Driving with llms: Fusing object-level vector modality for explainable autonomous driving},
  author={Chen, Long and Sinavski, Oleg and H{\"u}nermann, Jan and Karnsund, Alice and Willmott, Andrew James and Birch, Danny and Maund, Daniel and Shotton, Jamie},
  booktitle={ICRA},
  year={2024},
}

@article{cui2025drivemlm,
  author    = {Erfei Cui and Wenhai Wang and Zhiqi Li and Jiangwei Xie and Haoming Zou and Hanming Deng and Gen Luo and Lewei Lu and Xizhou Zhu and Jifeng Dai},
  title     = {DriveMLM: Aligning Multi-Modal Large Language Models with Behavioral Planning States for Autonomous Driving},
  journal   = {Volume 3},
  number    = {22},
  year      = {2025},
  url       = {https://link.springer.com/article//10.1007/s44267-025-00095-w}
}

@misc{china_2021_road_traffic_safety_law,
  author       = {Standing Committee of the National People's Congress of the People's Republic of China},
  title        = {Road Traffic Safety Law of the People's Republic of China (2021 Amendment)},
  howpublished = {\url{https://jtgl.beijing.gov.cn/jgj/jgxx/flfg/fl/205308/index.html}},
  year         = {2021},
  note         = {Originally adopted on Oct. 28, 2003; amended on Apr. 29, 2021; effective from May 1, 2024.},
  language     = {en}
}

@misc{china_2017_road_traffic_safety_law_regulations,
  author       = {State Council of the People's Republic of China},
  title        = {Regulations for the Implementation of the Road Traffic Safety Law of the People's Republic of China (2017 Amendment)},
  howpublished = {\url{https://www.gov.cn/gongbao/content/2019/content_5468932.htm}},
  year         = {2017},
  note         = {Originally adopted on April 28, 2004 (State Council Order No. 405); amended on Oct. 7, 2017 (State Council Order No. 687).},
  language     = {en}
}

@inproceedings{neuhold2017mapillary,
  title={The mapillary vistas dataset for semantic understanding of street scenes},
  author={Neuhold, Gerhard and Ollmann, Tobias and Rota Bulo, Samuel and Kontschieder, Peter},
  booktitle={ICCV},
  year={2017}
}

@inproceedings{marcu2024lingoqa,
  title={Lingoqa: Visual question answering for autonomous driving},
  author={Marcu, Ana-Maria and Chen, Long and H{\"u}nermann, Jan and Karnsund, Alice and Hanotte, Benoit and Chidananda, Prajwal and Nair, Saurabh and Badrinarayanan, Vijay and Kendall, Alex and Shotton, Jamie and others},
  booktitle={European Conference on Computer Vision},
  pages={252--269},
  year={2024},
  organization={Springer}
}

@misc{openai2025gpt5,
  title={GPT-5 System Card},
  author={OpenAI},
  year={2025},
  month={8},
  howpublished={OpenAI Website},
  note={https://cdn.openai.com/gpt-5-system-card.pdf}
}

@misc{anthropic2024Claude-Sonnet-4.5,
  title={System Card: Claude Sonnet 4.5},
  author={Anthropic},
  year={2025},
  month={9},
  howpublished={Anthropic Blog},
  note={https://www.anthropic.com/claude}
}

@misc{vteam2025glm45v,
      title={GLM-4.5V and GLM-4.1V-Thinking: Towards Versatile Multimodal Reasoning with Scalable Reinforcement Learning}, 
      author={V Team and Wenyi Hong and Wenmeng Yu and Xiaotao Gu and Guo Wang and Guobing Gan and Haomiao Tang and Jiale Cheng and Ji Qi and Junhui Ji and Lihang Pan and Shuaiqi Duan and Weihan Wang and Yan Wang and Yean Cheng and Zehai He and Zhe Su and Zhen Yang and Ziyang Pan and Aohan Zeng and Baoxu Wang and Bin Chen and Boyan Shi and Changyu Pang and Chenhui Zhang and Da Yin and Fan Yang and Guoqing Chen and Jiazheng Xu and Jiale Zhu and Jiali Chen and Jing Chen and Jinhao Chen and Jinghao Lin and Jinjiang Wang and Junjie Chen and Leqi Lei and Letian Gong and Leyi Pan and Mingdao Liu and Mingde Xu and Mingzhi Zhang and Qinkai Zheng and Sheng Yang and Shi Zhong and Shiyu Huang and Shuyuan Zhao and Siyan Xue and Shangqin Tu and Shengbiao Meng and Tianshu Zhang and Tianwei Luo and Tianxiang Hao and Tianyu Tong and Wenkai Li and Wei Jia and Xiao Liu and Xiaohan Zhang and Xin Lyu and Xinyue Fan and Xuancheng Huang and Yanling Wang and Yadong Xue and Yanfeng Wang and Yanzi Wang and Yifan An and Yifan Du and Yiming Shi and Yiheng Huang and Yilin Niu and Yuan Wang and Yuanchang Yue and Yuchen Li and Yutao Zhang and Yuting Wang and Yu Wang and Yuxuan Zhang and Zhao Xue and Zhenyu Hou and Zhengxiao Du and Zihan Wang and Peng Zhang and Debing Liu and Bin Xu and Juanzi Li and Minlie Huang and Yuxiao Dong and Jie Tang},
      year={2025},
      eprint={2507.01006},
      archivePrefix={arXiv},
      primaryClass={cs.CV},
      url={https://arxiv.org/abs/2507.01006}, 
}

@article{yang2025qwen3,
  title={Qwen3 technical report},
  author={Yang, An and Li, Anfeng and Yang, Baosong and Zhang, Beichen and Hui, Binyuan and Zheng, Bo and Yu, Bowen and Gao, Chang and Huang, Chengen and Lv, Chenxu and others},
  journal={arXiv preprint arXiv:2505.09388},
  year={2025}
}

@article{zhang2023Llama,
    author = {Zhang, Hang and Li, Xin and Bing, Lidong},
    journal = {arXiv preprint arXiv:2306.02858},
    title = {Video-llama: An Instruction-tuned Audio-visual Language Model for Video Understanding},
    year = {2023}
}

@misc{openScenario,
  title = {ASAM OpenSCENARIO: User Guide},
  howpublished = {\url{https://www.asam.net/index.php?eID=dumpFile&t=f&f=4092&token=d3b6a55e911b22179e3c0895fe2caae8f5492467}},
  year = "2021",
  author = "ASAM",
}

@article{team2025gemma3,
  title={Gemma 3 technical report},
  author={Team, Gemma and Kamath, Aishwarya and Ferret, Johan and Pathak, Shreya and Vieillard, Nino and Merhej, Ramona and Perrin, Sarah and Matejovicova, Tatiana and Ram{\'e}, Alexandre and Rivi{\`e}re, Morgane and others},
  journal={arXiv preprint arXiv:2503.19786},
  year={2025}
}

@article{song2025insertany,
  title={Insert anything: Image insertion via in-context editing in dit},
  author={Song, Wensong and Jiang, Hong and Yang, Zongxing and Quan, Ruijie and Yang, Yi},
  journal={arXiv preprint arXiv:2504.15009},
  year={2025}
}

@article{team2023gemini,
  title={Gemini: a family of highly capable multimodal models},
  author={Team, Gemini and Anil, Rohan and Borgeaud, Sebastian and Wu, Yonghui and Alayrac, Jean-Baptiste and Yu, Jiahui and Soricut, Radu and Schalkwyk, Johan and Dai, Andrew M and Hauth, Anja and others},
  journal={arXiv preprint arXiv:2312.11805},
  year={2023}
}

@article{lewis2020rag,
    author = {Lewis, Patrick and Perez, Ethan and Piktus, Aleksandra and Petroni, Fabio and Karpukhin, Vladimir and Goyal, Naman and K{\"u}ttler, Heinrich and Lewis, Mike and Yih, Wen-tau and Rockt{\"a}schel, Tim and others},
    journal = {NeurIPS},
    title = {Retrieval-augmented Generation for Knowledge-intensive Nlp Tasks},
    year = {2020}
}

@inproceedings{wang2025omnidrive,
  title={Omnidrive: A holistic vision-language dataset for autonomous driving with counterfactual reasoning},
  author={Wang, Shihao and Yu, Zhiding and Jiang, Xiaohui and Lan, Shiyi and Shi, Min and Chang, Nadine and Kautz, Jan and Li, Ying and Alvarez, Jose M},
  booktitle={Proceedings of the Computer Vision and Pattern Recognition Conference},
  pages={22442--22452},
  year={2025}
}

@article{wei2022cot,
    author = {Wei, Jason and Wang, Xuezhi and Schuurmans, Dale and Bosma, Maarten and Xia, Fei and Chi, Ed and Le, Quoc V and Zhou, Denny and others},
    journal = {NeurIPS},
    title = {Chain-of-thought Prompting Elicits Reasoning in Large Language Models},
    year = {2022}
}

@article{hu2022lora,
  title={Lora: Low-rank adaptation of large language models.},
  author={Hu, Edward J and Shen, Yelong and Wallis, Phillip and Allen-Zhu, Zeyuan and Li, Yuanzhi and Wang, Shean and Wang, Lu and Chen, Weizhu and others},
  journal={ICLR},
  volume={1},
  number={2},
  pages={3},
  year={2022}
}

@inproceedings{zheng2024llamafactory,
  title={LlamaFactory: Unified Efficient Fine-Tuning of 100+ Language Models},
  author={Yaowei Zheng and Richong Zhang and Junhao Zhang and Yanhan Ye and Zheyan Luo and Zhangchi Feng and Yongqiang Ma},
  booktitle={Proceedings of the 62nd Annual Meeting of the Association for Computational Linguistics (Volume 3: System Demonstrations)},
  address={Bangkok, Thailand},
  publisher={Association for Computational Linguistics},
  year={2024},
  url={http://arxiv.org/abs/2403.13372}
}

@inproceedings{renz2025simlingo,
  title={Simlingo: Vision-only closed-loop autonomous driving with language-action alignment},
  author={Renz, Katrin and Chen, Long and Arani, Elahe and Sinavski, Oleg},
  booktitle={Proceedings of the Computer Vision and Pattern Recognition Conference},
  pages={11993--12003},
  year={2025}
}

@inproceedings{park2024vlaad,
  title={Vlaad: Vision and language assistant for autonomous driving},
  author={Park, SungYeon and Lee, MinJae and Kang, JiHyuk and Choi, Hahyeon and Park, Yoonah and Cho, Juhwan and Lee, Adam and Kim, DongKyu},
  booktitle={Proceedings of the IEEE/CVF Winter Conference on Applications of Computer Vision},
  pages={980--987},
  year={2024}
}

@article{sima2023drivelm,
    author = {Sima, Chonghao and Renz, Katrin and Chitta, Kashyap and Chen, Li and Zhang, Hanxue and Xie, Chengen and Bei{\ss}wenger, Jens and Luo, Ping and Geiger, Andreas and Li, Hongyang},
    journal = {arXiv preprint arXiv:2312.14150},
    title = {Drivelm: Driving with Graph Visual Question Answering},
    year = {2023}
}

@inproceedings{xu2020BDDOIA,
  title={Explainable object-induced action decision for autonomous vehicles},
  author={Xu, Yiran and Yang, Xiaoyin and Gong, Lihang and Lin, Hsuan-Chu and Wu, Tz-Ying and Li, Yunsheng and Vasconcelos, Nuno},
  booktitle={Proceedings of the IEEE/CVF Conference on Computer Vision and Pattern Recognition},
  pages={9523--9532},
  year={2020}
}

@article{kim2018BDDX,
  title={Textual Explanations for Self-Driving Vehicles},
  author={Kim, Jinkyu and Rohrbach, Anna and Darrell, Trevor and Canny, John and Akata, Zeynep},
  journal={Proceedings of the European Conference on Computer Vision (ECCV)},
  year={2018}
}

@article{wilson2023argoverse,
    author = {Wilson, Benjamin and Qi, William and Agarwal, Tanmay and Lambert, John and Singh, Jagjeet and Khandelwal, Siddhesh and Pan, Bowen and Kumar, Ratnesh and Hartnett, Andrew and Pontes, Jhony Kaesemodel and others},
    journal = {arXiv preprint arXiv:2301.00493},
    title = {Argoverse 2: Next Generation Datasets for Self-driving Perception and Forecasting},
    year = {2023}
}

@inproceedings{sun2020Waymo,
    author = {Sun, Pei and Kretzschmar, Henrik and Dotiwalla, Xerxes and Chouard, Aurelien and Patnaik, Vijaysai and Tsui, Paul and Guo, James and Zhou, Yin and Chai, Yuning and Caine, Benjamin and others},
    booktitle = {CVPR},
    title = {Scalability in Perception for Autonomous Driving: Waymo Open Dataset},
    year = {2020}
}

@article{geiger2013KITTI,
  title={Vision meets robotics: The kitti dataset},
  author={Geiger, Andreas and Lenz, Philip and Stiller, Christoph and Urtasun, Raquel},
  journal={The international journal of robotics research},
  volume={32},
  number={11},
  pages={1231--1237},
  year={2013},
  publisher={Sage Publications Sage UK: London, England}
}

@article{mao2023agentdriver,
  title={A language agent for autonomous driving},
  author={Mao, Jiageng and Ye, Junjie and Qian, Yuxi and Pavone, Marco and Wang, Yue},
  journal={arXiv preprint arXiv:2311.10813},
  year={2023}
}

@article{hwang2024emma,
  title={Emma: End-to-end multimodal model for autonomous driving},
  author={Hwang, Jyh-Jing and Xu, Runsheng and Lin, Hubert and Hung, Wei-Chih and Ji, Jingwei and Choi, Kristy and Huang, Di and He, Tong and Covington, Paul and Sapp, Benjamin and others},
  journal={arXiv preprint arXiv:2410.23262},
  year={2024}
}

@inproceedings{li2022coda,
  title={Coda: A real-world road corner case dataset for object detection in autonomous driving},
  author={Li, Kaican and Chen, Kai and Wang, Haoyu and Hong, Lanqing and Ye, Chaoqiang and Han, Jianhua and Chen, Yukuai and Zhang, Wei and Xu, Chunjing and Yeung, Dit-Yan and others},
  booktitle={European conference on computer vision},
  pages={406--423},
  year={2022},
  organization={Springer}
}

@inproceedings{yu2020bdd100k,
  title={Bdd100k: A diverse driving dataset for heterogeneous multitask learning},
  author={Yu, Fisher and Chen, Haofeng and Wang, Xin and Xian, Wenqi and Chen, Yingying and Liu, Fangchen and Madhavan, Vashisht and Darrell, Trevor},
  booktitle={Proceedings of the IEEE/CVF conference on computer vision and pattern recognition},
  pages={2636--2645},
  year={2020}
}

@article{wang2024coda-lm,
  title={Rac3: Retrieval-augmented corner case comprehension for autonomous driving with vision-language models},
  author={Wang, Yujin and Liu, Quanfeng and Fan, Jiaqi and Hong, Jinlong and Chu, Hongqing and Tian, Mengjian and Gao, Bingzhao and Chen, Hong},
  journal={arXiv preprint arXiv:2412.11050},
  year={2024}
}

@article{xie2025DriveBench,
  title={Are vlms ready for autonomous driving? an empirical study from the reliability, data, and metric perspectives},
  author={Xie, Shaoyuan and Kong, Lingdong and Dong, Yuhao and Sima, Chonghao and Zhang, Wenwei and Chen, Qi Alfred and Liu, Ziwei and Pan, Liang},
  journal={arXiv preprint arXiv:2501.04003},
  year={2025}
}

@inproceedings{pan2024vlp,
  title={Vlp: Vision language planning for autonomous driving},
  author={Pan, Chenbin and Yaman, Burhaneddin and Nesti, Tommaso and Mallik, Abhirup and Allievi, Alessandro G and Velipasalar, Senem and Ren, Liu},
  booktitle={Proceedings of the IEEE/CVF Conference on Computer Vision and Pattern Recognition},
  pages={14760--14769},
  year={2024}
}

@inproceedings{malla2023drama,
  title={Drama: Joint risk localization and captioning in driving},
  author={Malla, Srikanth and Choi, Chiho and Dwivedi, Isht and Choi, Joon Hee and Li, Jiachen},
  booktitle={Proceedings of the IEEE/CVF winter conference on applications of computer vision},
  pages={1043--1052},
  year={2023}
}

@article{deruyttere2019talk2car,
  title={Talk2car: Taking control of your self-driving car},
  author={Deruyttere, Thierry and Vandenhende, Simon and Grujicic, Dusan and Van Gool, Luc and Moens, Marie-Francine},
  journal={arXiv preprint arXiv:1909.10838},
  year={2019}
}

@inproceedings{qian2024nuscenes-QA,
    author = {Qian, Tianwen and Chen, Jingjing and Zhuo, Linhai and Jiao, Yang and Jiang, Yu-Gang},
    booktitle = {AAAI},
    title = {Nuscenes-qa: A Multi-modal Visual Question Answering Benchmark for Autonomous Driving Scenario},
    year = {2024}
}

@inproceedings{caesar2020nuscenes,
    author = {Caesar, Holger and Bankiti, Varun and Lang, Alex H and Vora, Sourabh and Liong, Venice Erin and Xu, Qiang and Krishnan, Anush and Pan, Yu and Baldan, Giancarlo and Beijbom, Oscar},
    booktitle = {CVPR},
    title = {Nuscenes: A Multimodal Dataset for Autonomous Driving},
    year = {2020}
}

@inproceedings{lu2025idkb,
  title={Can lvlms obtain a driver’s license? a benchmark towards reliable agi for autonomous driving},
  author={Lu, Yuhang and Yao, Yichen and Tu, Jiadong and Shao, Jiangnan and Ma, Yuexin and Zhu, Xinge},
  booktitle={Proceedings of the AAAI Conference on Artificial Intelligence},
  volume={39},
  number={6},
  pages={5838--5846},
  year={2025}
}

@article{tian2024drivevlm,
    author = {Tian, Xiaoyu and Gu, Junru and Li, Bailin and Liu, Yicheng and Wang, Yang and Zhao, Zhiyong and Zhan, Kun and Jia, Peng and Lang, Xianpeng and Zhao, Hang},
    journal = {arXiv preprint arXiv:2402.12289},
    title = {Drivevlm: The Convergence of Autonomous Driving and Large Vision-language Models},
    year = {2024}
}

@article{xu2024drivegpt4,
    author = {Xu, Zhenhua and Zhang, Yujia and Xie, Enze and Zhao, Zhen and Guo, Yong and Wong, Kwan-Yee K and Li, Zhenguo and Zhao, Hengshuang},
    journal = {RA-L},
    title = {Drivegpt4: Interpretable End-to-end Autonomous Driving Via Large Language Model},
    year = {2024}
}

@article{deng2025target,
  title={Target: Traffic rule-based test generation for autonomous driving systems},
  author={Deng, Yao and Tu, Zhi and Yao, Jiaohong and Zhang, Mengshi and Zhang, Tianyi and Zheng, Xi},
  journal={IEEE Transactions on Software Engineering},
  year={2025},
  publisher={IEEE}
}

@inproceedings{dosovitskiy2017carla,
  title={CARLA: An open urban driving simulator},
  author={Dosovitskiy, Alexey and Ros, German and Codevilla, Felipe and Lopez, Antonio and Koltun, Vladlen},
  booktitle={Conference on robot learning},
  pages={1--16},
  year={2017},
  organization={PMLR}
}

@inproceedings{wei2025driveqa,
  title={Passing the driving knowledge test},
  author={Wei, Maolin and Liu, Wanzhou and Ohn-Bar, Eshed},
  booktitle={Proceedings of the IEEE/CVF International Conference on Computer Vision},
  pages={8395--8406},
  year={2025}
}
}

\clearpage
\setcounter{page}{1}
\maketitlesupplementary

In this supplemental material, we provide additional details of our proposed DriveCombo benchmark. The content is organized as follows:

\begin{itemize}
    \item \textbf{Section~\ref{sec:detail_dataset}} provides additional details on the construction of the DriveCombo dataset.

    \item \textbf{Section~\ref{sec:detail_Experimentsetup}} explains further experimental settings for evaluating MLLMs on DriveCombo.

    \item \textbf{Section~\ref{sec:Analysis_Results}} presents more analyses of the evaluation results, covering multi–rule compositional reasoning capabilities, and the potential of DriveCombo for closed-loop planning evaluation.
\end{itemize}

\section{More Details about datasets}
\label{sec:detail_dataset}
\subsection{Candidate Rule Pairs Generation for Speed-Limit Rules}
Speed-limit regulations exhibit numerical constraints that require a specialized pairing mechanism beyond the general normative-relation formulation described in the main paper. To properly model semantic and normative relationships among speed-related rules, we introduce a rule-pair generation strategy based on \emph{speed-range intersection}.

\paragraph{Action-Type Filtering.}
We first isolate all atomic rules whose action type corresponds to speed regulation. \(P = \left\{ p_i = \{ r_j, r_k \} \mid a_j = a_k = \text{SpeedLimit},\ r_j, r_k \in R \right\}_{i=1}^{|P|}.\)

\paragraph{Speed-Range Modeling.}
Each speed-limit rule is mapped to a feasible speed interval: \(sr_j = [l_j,\, u_j]\) and \(sr_k = [l_k,\, u_k]\) where $l_*$ and $u_*$ denote the lower and upper bounds extracted from the rule's semantic content (e.g., ``max 70 km/h,'' ``min 110 km/h on expressways,'' or ``max 30 km/h when visibility $<$ 50 m'').

\paragraph{Derived Labels.}
For each candidate pair $p_i = \{r_j, r_k\}$, we assign three derived labels following the structure of the hierarchical rule system.

Each pair \( p_i \) is assigned three derived labels: the perceptual combination type ${b'_i}$, the normative relation type of speed \( ns'_i \), and the hierarchical level \( l_i \):  
\[
b_i' =
\begin{cases}
\text{Double Static}, & b_j=b_k=\text{static},\\
\text{Double Dynamic}, & b_j=b_k=\text{dynamic},\\
\text{Hybrid}, & \text{otherwise.}
\end{cases}
\]
\[
ns'_i =
\begin{cases}
\text{Norm Conflict}, & sr_j \cap sr_k = \emptyset, \\
\text{Norm Harmony}, & sr_j \cap sr_k \neq \emptyset.
\end{cases}
\]
\[
l_i =
\begin{cases}
2, & b_i'=\text{Double Static}, n_i'=\text{Norm Harmony},\\
3, & b_i'=\text{Double Dynamic}, n_i'=\text{Norm Harmony},\\
4, & b_i'=\text{Hybrid}, n_i'=\text{Norm Harmony},\\
5, & n_i'=\text{Norm Conflict}.
\end{cases}
\]

\noindent
Notably, all conflicting speed-limit rule pairs (\,$sr_j \cap sr_k = \emptyset$\,) are directly assigned to Level~5, as numerical incompatibility reflects a genuine priority-arbitration scenario in real driving.

\subsection{Construction of Multi-level Questions}
We create multi-level multiple-choice questions (L1–L5). Each includes a scenario, a standardized stem (e.g., “What should the driver do in this situation?”), and four options.

\paragraph{Correct Action Determination.}  
The correct action \( a_i \) is generated according to the hierarchical rule level to which it belongs. For single-rule questions (L1), the correct action is directly specified by the atomic rule \( r_i \). For compatible rule levels (L2–L4), \( a_i \) represents a behavior that simultaneously satisfies the semantic and normative constraints of the rule pair. When rule pairs at the conflict level (L5) yield contradictory instructions, \( a_i \) is determined according to the following priority principle~\cite{china_2021_road_traffic_safety_law, china_2017_road_traffic_safety_law_regulations}. This principle ensures consistent and rational driving behavior in multi-signal conflicts, aligning the model’s decisions with ethical and legal standards when normative cues compete. The behavioral priority order is summarized as follows:
\begin{quote}
\textit{Pedestrian safety $>$ emergency vehicle avoidance $>$ on-site command $>$ traffic lights $>$ traffic signs $>$ road markings $>$ interactive right-of-way $>$ defensive driving $>$ emergency exceptions.}  
\end{quote}

\paragraph{Option Design.}  
Each question includes four options  
\( o_i = \{ o_i^*, o_{i1}, o_{i2}, o_{i3} \} \),  
where \( o_i^* \) is the correct, norm-compliant option derived from the correct action \( a_i \).  
The other three distractors are generated by the LLM and manually filtered to ensure semantic plausibility while introducing normative deviations, including priority confusion, minor violations, or incorrect yielding. All options are jointly verified by the LLM and human experts

Finally, the final question set \( T = \{ (q_i, i_i, o_i) \}_{i=1}^{|S|} \) is ready,  
where \( q_i \) denotes the question stem, \( i_i \) the sequence of visual input frames, \( o_i \) the option set. 

\paragraph{Specific Generation Process.}  
For each rule or rule pair, we first prompt Claude~Sonnet~4.5~\cite{anthropic2024Claude-Sonnet-4.5} to generate a full MCQ sample in a fixed JSON format containing a scenario description, question stem, four options (A--D), the design logic, the correct answer, and an explanation. The prompt is shown in Table~\ref{tab:prompt_of_mcq_generation_1}, \ref{tab:prompt_of_mcq_generation_2_3_4} and
\ref{tab:prompt_of_mcq_generation_5}. Next, to ensure correctness and rule consistency, each generated question is independently validated by Gemini~2.5~Pro~\cite{team2023gemini}, GPT~5~Pro~\cite{openai2025gpt5}, and Claude~Sonnet~4.5~\cite{anthropic2024Claude-Sonnet-4.5}, which assess whether the answer is logically valid, whether the scenario faithfully reflects the input rules, and whether the options are well-structured. The prompt is shown in Table~\ref{tab:prompt_of_mcq_generation_check}. Each model outputs a binary decision (\texttt{1} for valid, \texttt{0} for invalid), and a question is accepted only if all three models return \texttt{1}; otherwise, it is regenerated. After automated filtering, 5\% of the accepted questions are randomly sampled for human expert verification to ensure overall quality and alignment with the intended cognitive levels.

\subsection{Multi-Country Rule and Jurisdiction Modeling}
To account for cross-country rule differences (e.g., driving side) and evaluate MLLM adaptability across regulatory systems, DriveCombo avoids mixing traffic rules across countries within the same evaluation instance. Instead, separate rule sets are constructed for each country, and modeling and evaluation are conducted independently. All countries’ rules are built using the same data processing pipeline to ensure consistent rule modeling. We further report per-country performance in Sec.~\ref{sec:supp_multi_country_results}. In addition, the jurisdiction identity is explicitly specified in each question prompt, ensuring that MLLMs reason under the correct country-specific traffic rules.

\subsection{Prompt Details for Scene Generation}
This section provides detailed descriptions of the prompt designs used in the Rule2Scene pipeline, including semantic structuring, coexistence validation, scenario transcription, DSL translation with self-consistency checking, and LLM-based quality scoring. 

\paragraph{Semantic Structuring Prompt.}
In the semantic structuring stage, the LLM is instructed to transform a natural-language traffic rule into a normalized atomic representation containing rule content, perceptual types, norm types, action types, and numeric constraints. This step ensures that every rule is converted into a consistent machine-interpretable structure before rule pairing or scene generation. The prompt is shown in Table~\ref{tab:prompt-semantic-structuring}.

\paragraph{Coexistence Validation Prompt.}
To determine whether the given atomic rules can coexist in a physically and semantically coherent driving scenario, the LLM receives a paired-rule input and evaluates spatial compatibility, temporal feasibility, weather or road-type requirements, and agent-role consistency. The model outputs a binary indicator representing feasibility. Only rule pairs that are jointly executable proceed to scenario generation. This filtering step prevents contradictory or impossible combinations from entering higher-level compositional reasoning tasks. The prompt is shown in Table~\ref{tab:prompt_combination_check}.

\paragraph{Scenario Transcription Prompt.}
For each validated rule or rule pair, the LLM is prompted to convert the structured rule representation into a high-level natural-language scene description. This transcription step bridges abstract normative semantics with concrete driving contexts and provides the narrative foundation for subsequent DSL conversion and visual rendering. The prompt instructs the model to produce detailed but concise scene descriptions that correctly embed all rule constraints. This process is embedded within the prompt used for problem generation.

\paragraph{DSL Translation Prompt.}
After producing the scene description, the LLM is prompted to translate the textual scenario into a structured semantic DSL representation, including entities, relations, positions, trajectories, and environmental settings. The prompt is shown in Table~\ref{tab:prompt-dsl-translation}.

\paragraph{LLM-Based Quality Scoring.}
To ensure the reliability of the generated scenarios, we incorporate an LLM-based quality assessment mechanism across all four stages of the pipeline: \textit{Semantic Structuring}, \textit{Coexistence Validation}, \textit{Scenario Transcription}, and \textit{DSL Translation}. At each stage, three advanced language models, Gemini~2.5~Pro~\cite{team2023gemini}, GPT~5~Pro~\cite{openai2025gpt5}, and Claude~Sonnet~4.5~\cite{anthropic2024Claude-Sonnet-4.5}, independently evaluate the generated output and assign a quality score within the range $[0, 1]$. The final quality score is computed as the arithmetic mean of the three model predictions. If the average score falls below $0.6$, the system automatically flags the sample and requests human intervention to ensure correctness and consistency. The prompt is shown in Table~\ref{tab:prompt-quality-scoring-semantic}, \ref{tab:prompt-quality-scoring-coexistence}, \ref{tab:prompt-quality-scoring-transcription} and \ref{tab:prompt-quality-scoring-dsl}.

\subsection{Action Type Set}

To comprehensively represent the behavioral space of real-world driving, DriveCombo defines a structured action type set \(\mathcal{A}\) composed of three major categories: \textit{Driving Maneuvers}, \textit{Lighting \& Signaling}, and \textit{Parking \& Yielding}. These categories cover the full range of decision-making primitives required for traffic-rule reasoning and scenario generation.

\paragraph{Driving Maneuvers.}
This category captures fundamental vehicle motion behaviors and lane-level interactions.
\begin{itemize}
    \item \textit{Overtake}: Move ahead of a slower vehicle by passing it safely.
    \item \textit{Left Turn}: Steer the vehicle to enter a road or lane on the left side.
    \item \textit{Right Turn}: Steer the vehicle to enter a road or lane on the right side.
    \item \textit{U-turn}: Rotate the vehicle 180 degrees to reverse its direction of travel.
    \item \textit{Lane Change}: Move laterally from one lane to another while maintaining direction.
    \item \textit{Merge Main Road}: Enter a primary roadway from a branch or side lane.
    \item \textit{Enter Ramp}: Access a highway or exit using an on-ramp or off-ramp.
    \item \textit{Acceleration}: Increase the vehicle's speed to match traffic flow or complete maneuvers.
    \item \textit{Deceleration}: Reduce speed to adapt to road conditions or prepare for stopping.
    \item \textit{Reverse}: Move the vehicle backward using rearward motion.
    \item \textit{Emergency Lane Usage}: Drive or stop on the emergency lane under special conditions.
\end{itemize}

\paragraph{Lighting and Signaling.}
This category includes communication and visibility-related actions that convey driver intent.
\begin{itemize}
    \item \textit{Left Turn Signal}: Activate the left indicator to show intention to turn or change lanes leftward.
    \item \textit{Right Turn Signal}: Activate the right indicator to signal rightward turning or lane change.
    \item \textit{Low Beam}: Use dipped headlights for normal nighttime or low-light driving.
    \item \textit{High Beam}: Use strong headlights to extend visibility when no oncoming traffic is present.
    \item \textit{Flashing Headlights}: Briefly flash headlights to warn or signal other road users.
    \item \textit{Double Flashers}: Activate hazard lights to indicate emergencies or temporary stopping.
    \item \textit{Fog Lights}: Use specialized lights designed for low-visibility foggy conditions.
    \item \textit{Position Lights}: Turn on minimal lighting to indicate the vehicle's presence when stationary or parked.
    \item \textit{Honk Horn}: Use the horn to warn pedestrians or communicate urgency to nearby vehicles.
\end{itemize}

\paragraph{Parking and Yielding.}
This category covers controlled stop-and-yield maneuvers essential for compliance with road regulations.
\begin{itemize}
    \item \textit{Temporary Parking}: Stop the vehicle briefly at a roadside location without turning off the engine.
    \item \textit{Pull Over}: Move the vehicle to the roadside for inspection, hazard avoidance, or instructions.
    \item \textit{Yield}: Give way to pedestrians or other vehicles according to right-of-way rules.
\end{itemize}

Together, these action types define the complete action space \(\mathcal{A}\) used throughout DriveCombo for atomic rule extraction, rule pairing, and scene generation. The coverage of lane operations, signaling behaviors, and yielding-related actions ensures that the benchmark faithfully reflects the breadth of real-world driving rules and constraints.

\subsection{Scene Mapping in Simulator}
\begin{figure*}
    \centering
      \includegraphics[width=0.93\linewidth]{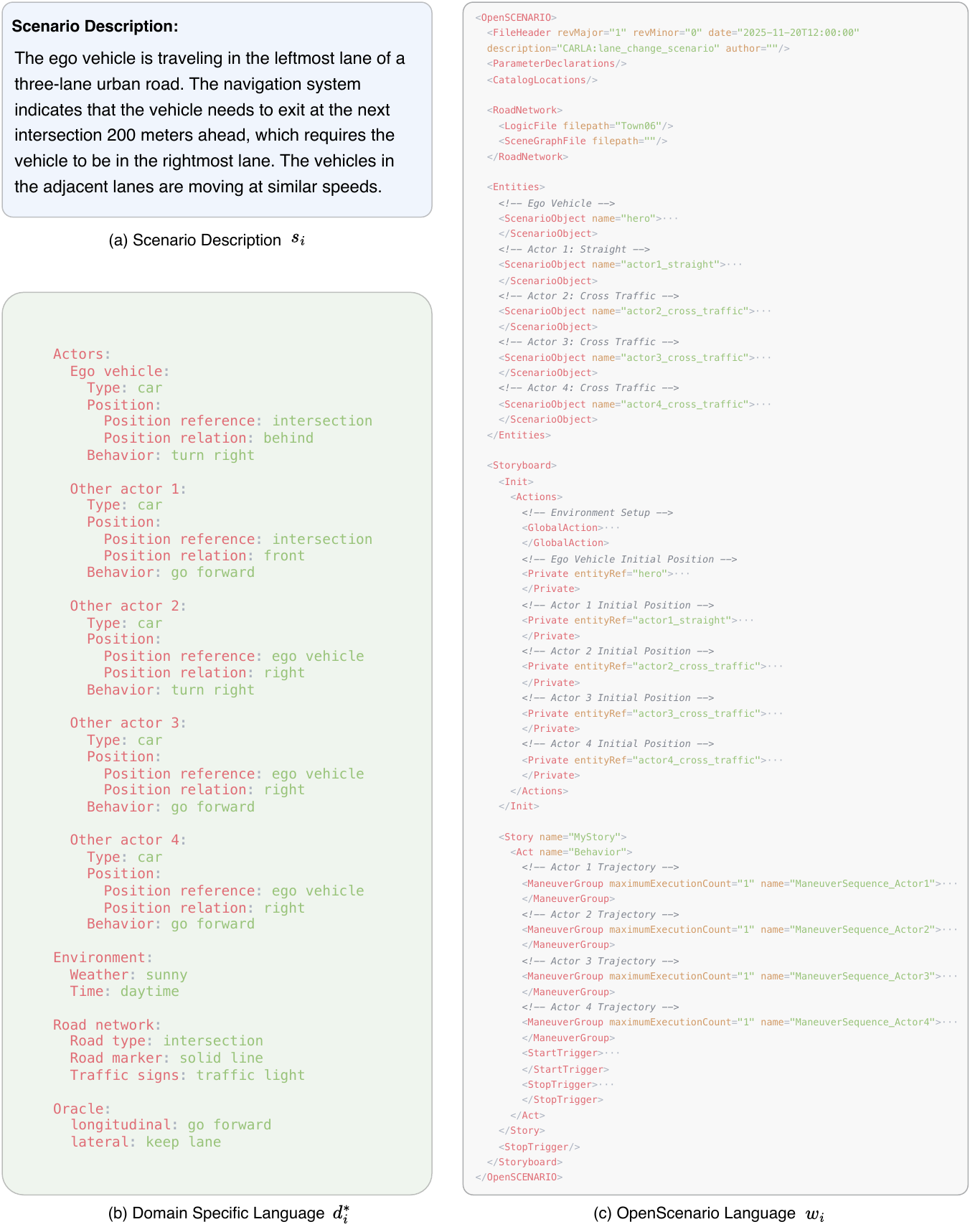}
    \caption{\textbf{Illustration of the three intermediate outputs of Scene Weaver in our proposed Rule2Scene Agent}. First, the input rules are converted into (a) a complex driving scenario $s_i$, which is then transformed into (b) a Domain Specific Language $d_i^*$. Finally, the scenario is mapped into (c) an OpenSCENARIO language $w_i$ that can be executed in the CARLA simulator.
    }
    \label{fig: supp_txt_yaml_xosc} 
\end{figure*}

The pipeline takes structured semantics $d_i^*$ as input and converts the abstract semantics to a standardized OpenSCENARIO~\cite{openScenario} dynamic scene  $w_i$ through layered parsing, instantiation, and trajectory generation. 
To make the pipeline clearer, we illustrate the three intermediate outputs of Scene Weaver in our proposed Rule2Scene Agent in Figure~\ref{fig: supp_txt_yaml_xosc}.
It is worth noting that our pipeline is built on TARGET~\cite{deng2025target} and further extends it from synthesizing scenes that support only a single atomic traffic rule to synthesizing complex scenes that support multiple traffic rules. We also expand the capability from inserting only one actor to inserting multiple actors, and introduce a new ability for joint inference of dynamic trajectories for multiple agents.

\paragraph{Construction of Static Initial Scene.}
This stage transforms the semantic description into an initial scene state with explicit spatial and environmental attributes. This stage first maps the abstract categories to the assets that can be loaded in CARLA~\cite{dosovitskiy2017carla} according to the entity type configuration in the structured semantics $d_i^*$. It then obtains precise geometric information through CARLA road topology queries, lane centerline extraction, and coordinate transformation interfaces. This information is used to compute the initial position, orientation, and speed of each traffic participant, while also supporting relative constraint-based generation and region-specified sampling. In addition, this stage parses the environmental conditions based on the weather field in $d_i^*$. It applies rule-based mappings to convert weather semantics in $d_i^*$ into concrete weather parameters required by CARLA to ensure executability during simulation. To ensure overall scene usability, the process continuously performs positional legality checks, collision detection, orientation validation, and boundary condition verification, and resamples when necessary. The static scene produced at this stage contains complete initial entity states and environmental settings, providing the static initial scene for subsequent trajectory generation.

\paragraph{Trajectory Generation and Dynamic Scene Construction.}
This stage generates complete temporal trajectories based on the static initial state described above, enabling continuous dynamic evolution of the scene in CARLA. The pipeline first generates feasible paths for each entity according to CARLA road topology and lane connectivity, while inferring driving strategies from the behavioral semantics in $d_i^*$. The strategies include driving along the centerline, lane changes, following, interactive approaching, and pedestrian navigation. After generating the paths, the pipeline samples the trajectories in the temporal dimension. These temporal trajectories are then encoded as OpenSCENARIO action structures and rewritten into motion commands that can be executed directly in CARLA through the trajectory element, as shown in Figure~\ref{fig: supp_txt_yaml_xosc}(c). This stage integrates spatial layout with traffic behaviors, producing motion that is physically plausible and consistent with the topology.

Through integration with CARLA map topology, the pipeline establishes a continuous and coherent transformation pipeline among semantic parsing, spatial instantiation, and temporal trajectory generation, thereby achieving a complete generation process from high-level intent to executable traffic simulation.

\subsection{More task examples of DriveCombo}
\begin{figure*}
    \centering
      \includegraphics[width=0.75\linewidth]{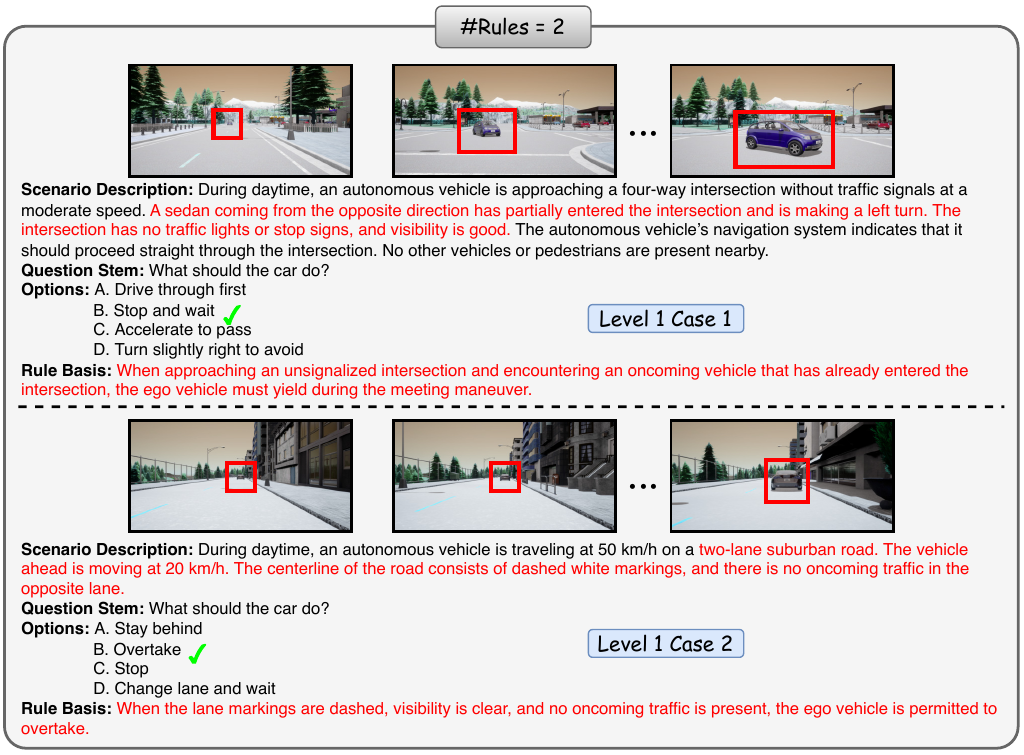}
    \caption{ \textbf{More Examples of Level 1 task in DriveCombo under \#Rules=2 setting}.
    }
    \label{fig: com2_sup_level_1}
\end{figure*}

\begin{figure*}
    \centering
      \includegraphics[width=0.75\linewidth]{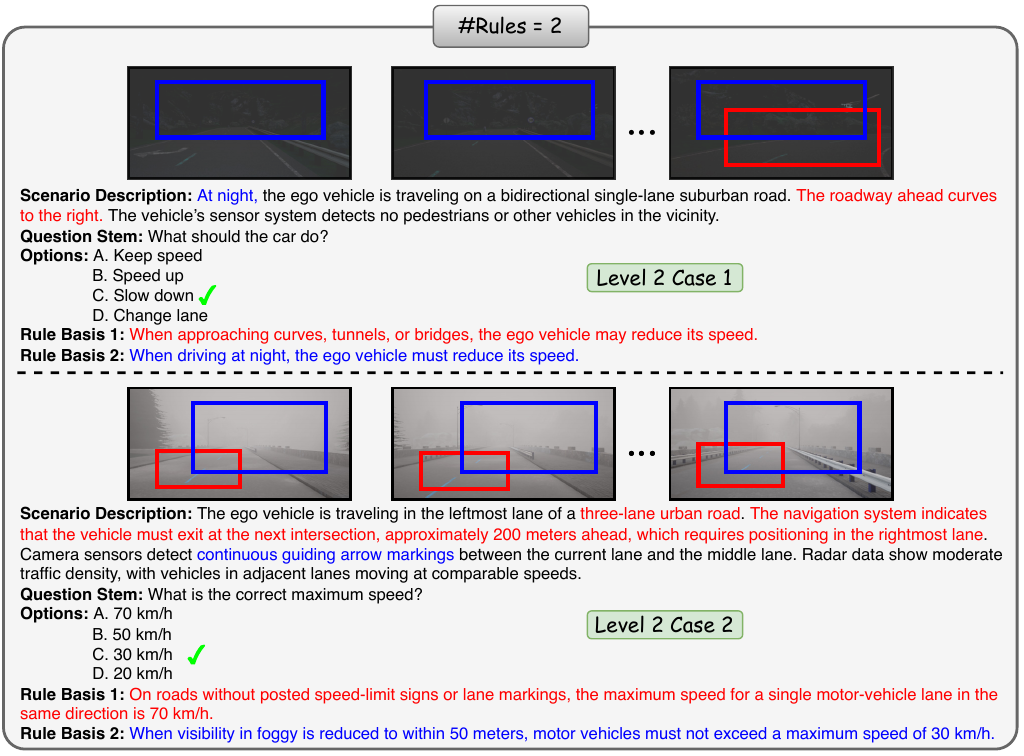}
    \caption{\textbf{More Examples of Level 2 task in DriveCombo under \#Rules=2 setting}.
    }
    \label{fig: com2_sup_level_2} 
\end{figure*}

\begin{figure*}
    \centering
      \includegraphics[width=0.75\linewidth]{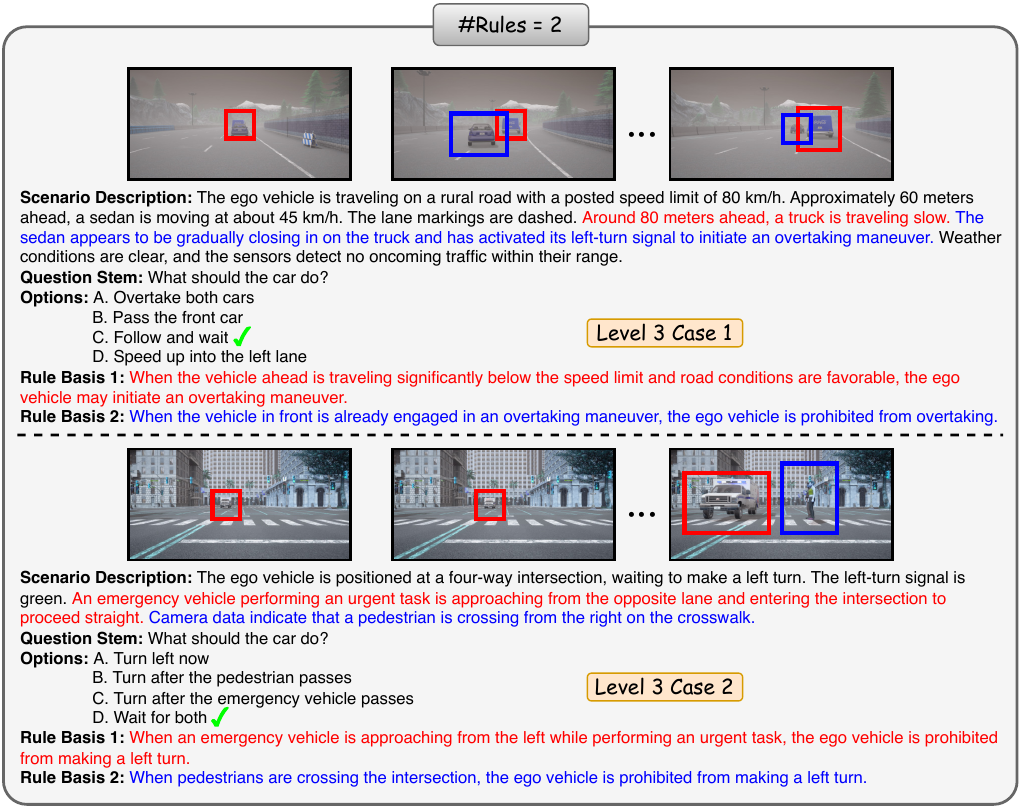}
    \caption{ \textbf{More Examples of Level 3 task in DriveCombo under \#Rules=2 setting}.
    }
    \label{fig: com2_sup_level_3}
\end{figure*}

\begin{figure*}
    \centering
      \includegraphics[width=0.75\linewidth]{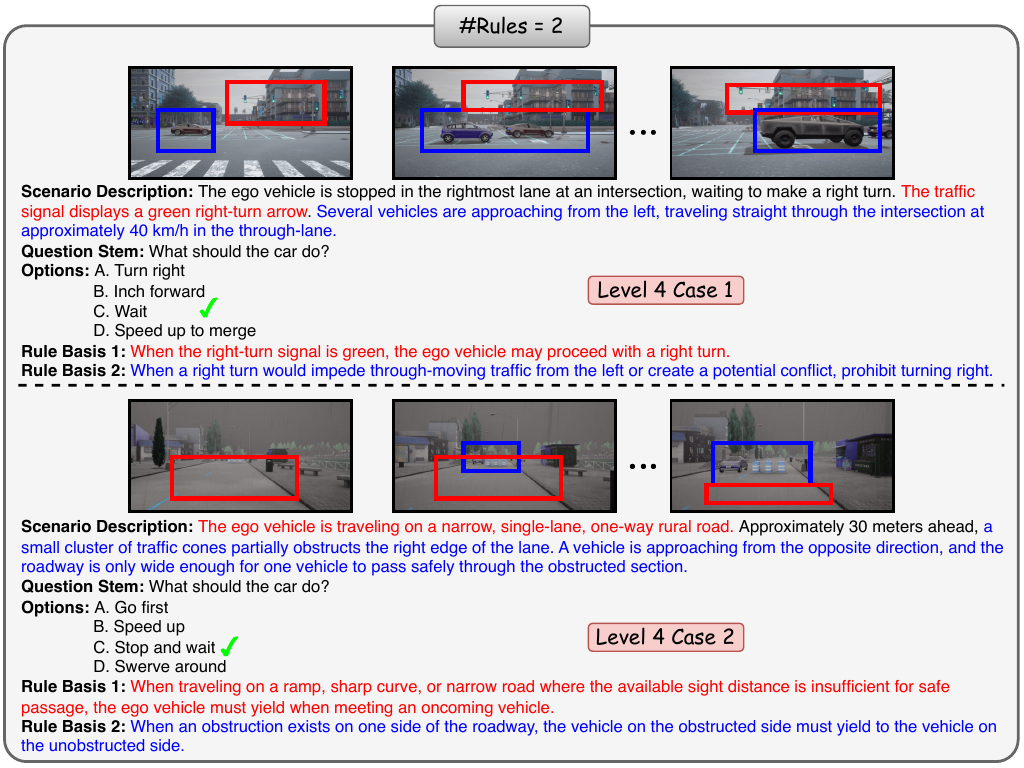}
    \caption{ \textbf{More Examples of Level 4 task in DriveCombo under \#Rules=2 setting}.
    }
    \label{fig: com2_sup_level_4} 
\end{figure*}

\begin{figure*}
    \centering
      \includegraphics[width=0.75\linewidth]{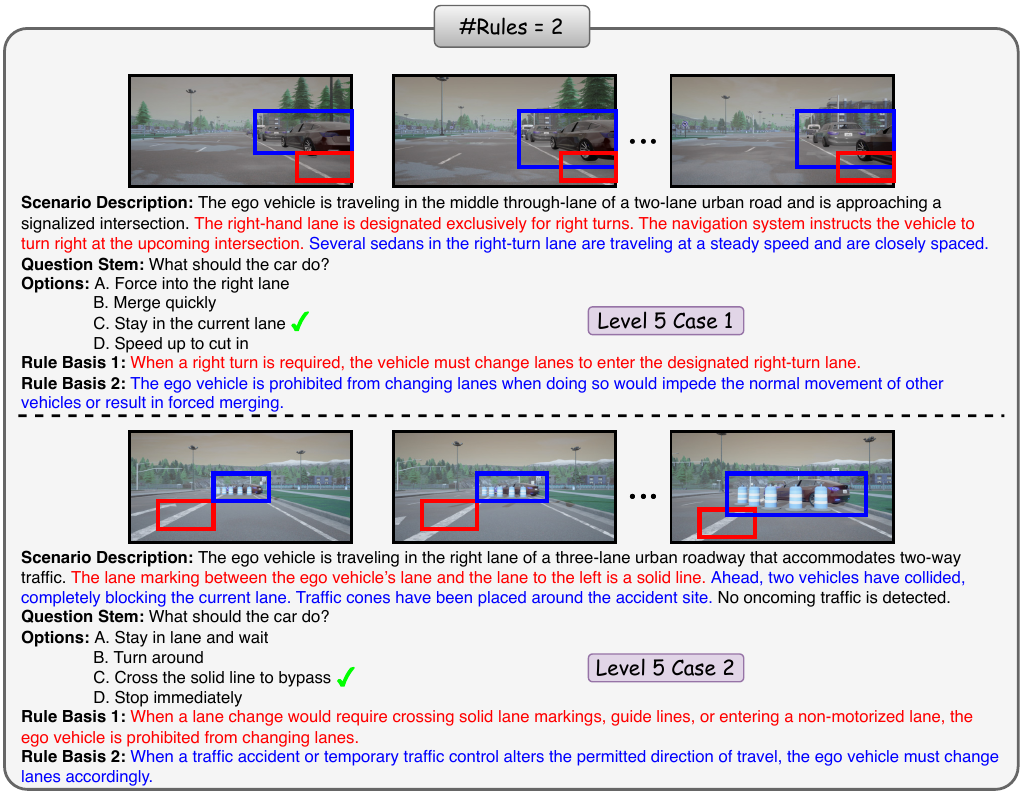}
    \caption{ \textbf{More Examples of Level 5 task in DriveCombo under \#Rules=2 setting}.
    }
    \label{fig: com2_sup_level_5} 
\end{figure*}

\begin{figure*}
    \centering
      \includegraphics[width=0.9\linewidth]{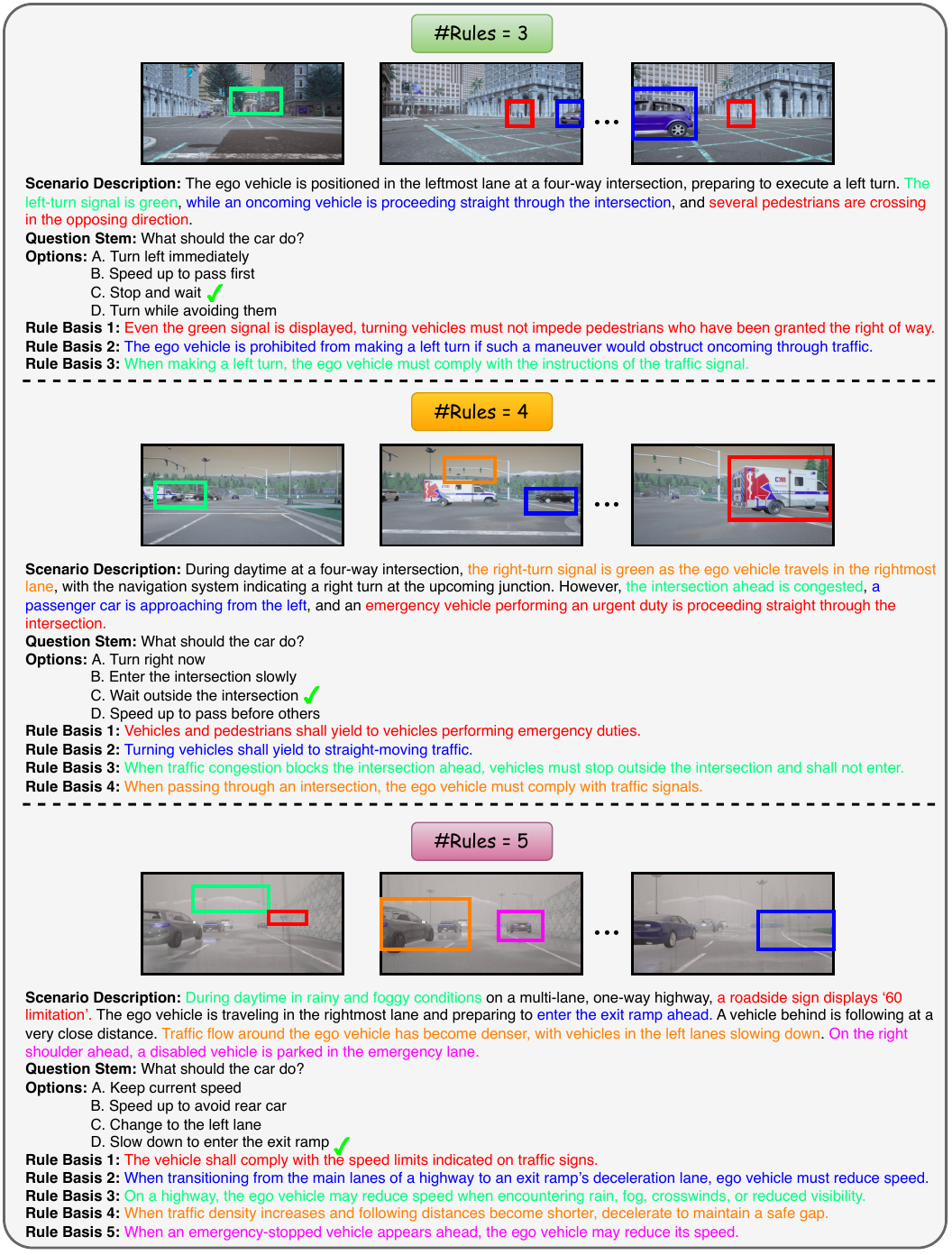}
    \caption{ \textbf{More Examples of DriveCombo under \#Rules=3, \#Rules=4, and \#Rules=5 setting}.
    }
    \label{fig: sup_comb_3_4_5} 
\end{figure*}

We provide additional examples of MCQs under the \#Rules=2 setting in DriveCombo, with Figures~\ref{fig: com2_sup_level_1} through \ref{fig: com2_sup_level_5} corresponding to tasks L1 through L5. 
In addition, we present examples involving combinations of multiple traffic rules, with Figure~\ref{fig: sup_comb_3_4_5} showing MCQs examples under the \#Rules=3, \#Rules=4, and \#Rules=5 settings.

\section{More Details about Experiment Setup}
\label{sec:detail_Experimentsetup}
\subsection{General Evaluation Setup}
To ensure reproducibility and eliminate variance introduced by stochastic decoding, all models are evaluated using a deterministic greedy decoding strategy. Specifically, we set temperature to 0 and fix both top-p and top-k to 1 for all inference runs.

For the DriveCombo benchmark, the evaluation input consists of four visual frames extracted from each generated scenario, serving as the multimodal observation for the model. For the DriveCombo-Text variant, we replace visual inputs with the corresponding textual scene descriptions while keeping all other components identical. The prompt is shown in Table~\ref{tab:prompt-test} and \ref{tab:prompt-test-tqa}.

\subsection{Enhancement Evaluation Setup}
\label{sec:enhancement}

To further investigate whether conventional reasoning-enhancement strategies can mitigate the performance gaps observed in DriveCombo, we evaluate three representative approaches: Chain-of-Thought (CoT)~\cite{wei2022cot}, Retrieval-Augmented Generation (RAG)~\cite{lewis2020rag}, and Supervised Fine-Tuning (SFT). The following details the implementation of each method.

\paragraph{Chain-of-Thought (CoT).}
For CoT prompting, we apply standard step-by-step reasoning instructions without additional supervision or fine-tuning. During inference, each model receives a fixed CoT template encouraging explicit logical decomposition (``think step by step''), followed by answering the final multiple-choice question. No rule-specific optimization or task-dependent heuristics are applied, ensuring a training-free enhancement protocol.

\paragraph{Retrieval-Augmented Generation (RAG).}
For RAG, we construct a retrieval corpus consisting exclusively of the \emph{original traffic rule books} used in building our atomic rule set $R$. Each model retrieves top-$k$ relevant passages (with $k=5$) using a dense retriever and conditions its answer on the concatenation of the query, retrieved rule excerpts, and the scenario description. This design ensures that the retrieved content reflects authentic legal constraints and avoids contamination from model-generated text. No scenario frames or DriveCombo samples are included in the retrieval database.

\paragraph{Supervised Fine-Tuning (SFT).}
For supervised adaptation, we employ the LLaMA-Factory framework~\cite{zheng2024llamafactory} to fine-tune a set of representative open-source models: Gemma~3~4B~\cite{team2025gemma3}, Gemma~3~12B, Qwen3-VL~2B~\cite{yang2025qwen3}, Qwen3-VL~8B, and Llama~3.2~11B~\cite{zhang2023Llama}. Our goal is to enhance the models' capability for compositional traffic-rule reasoning through direct exposure to DriveCombo training samples. All experiments are conducted on a cluster equipped with 8 NVIDIA H800 GPUs (80GB per GPU). We adopt LoRA (Low-Rank Adaptation)~\cite{hu2022lora} as the core optimization strategy. By leveraging low-rank matrix decomposition, LoRA enables efficient parameter adjustment while avoiding the substantial computational cost associated with full fine-tuning. Specifically, we set the LoRA rank to 8, an empirically determined configuration that balances model expressiveness, the volume of parameter updates, and memory consumption. To ensure the efficiency and stability of distributed training, we employ the DeepSpeed ZeRO-3 optimization strategy. This approach partitions model parameters, gradients, and optimizer states, thereby enabling highly efficient memory management and computational scheduling. As a result, it significantly improves training speed and parallelism in multi-GPU or multi-node environments. Regarding training configuration, we set the per-device batch size to 4, which helps avoid out-of-memory errors under limited GPU memory while maintaining a reasonable level of diversity. We further set the gradient accumulation steps to 4, allowing us to simulate a larger effective batch size despite the small actual batch size. This improves the accuracy of gradient estimation without increasing the computational load per step. To facilitate stable convergence, we use a learning rate of 1e-4. Finally, we train the model for three epochs, striking a balance between computational efficiency and adequate model refinement under resource constraints.

\section{More Analysis of Results}
\label{sec:Analysis_Results}

\begin{figure*}
    \centering
      \includegraphics[width=0.95\linewidth]{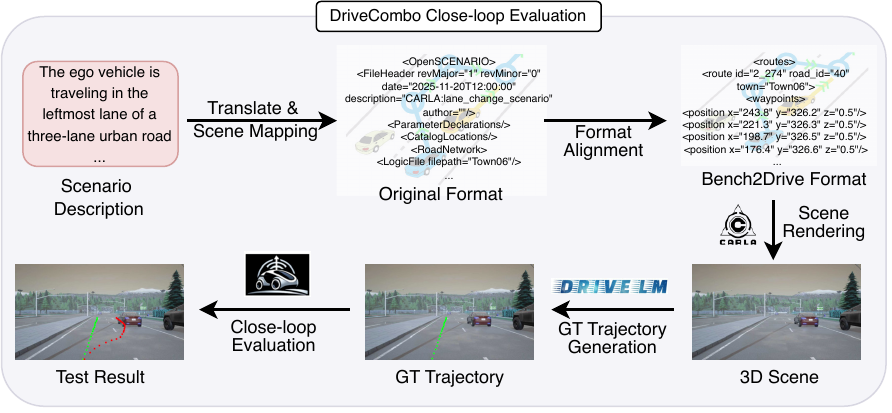}
    \caption{\textbf{DriveCombo’s scenarios can be seamlessly converted into closed-loop test cases to evaluate end-to-end models under real rule constraints.} The green line means the GT trajectory and the red means the test trajectory.
    }
    \label{fig: supp_close_loop} 
\end{figure*}

\begin{table*}
\centering
\caption{\textbf{Closed-loop Results of E2E-AD Methods in Bench2Drive and our DriveCombo.}}
\label{tab:supp_eval_on_bench2drive}
\begin{tabular}{l|c|cccc}
\toprule
\multirow{2}{*}{\textbf{Method}}     & \multirow{2}{*}{\textbf{Benchmark}} & \multicolumn{4}{c}{\textbf{Closed-loop Metrics}}                              \\
                            &                            & \textbf{Driving Score} ↑ & \textbf{Success Rate} (\%) ↑ & \textbf{Efficiency} ↑ & \textbf{Comfortness} ↑ \\
\midrule
\multirow{2}{*}{UniAD-Base~\cite{hu2023uniad}} & Bench2Drive~\cite{jia2024bench2drive}                & {45.81}           & {16.36}               & {129.21}       & {43.58}         \\
                            & DriveCombo                   & 27.59           & 13.33                & 101.32        & 31.92         \\
\midrule
\multirow{2}{*}{VAD~\cite{Jiang2023VADVS}}        & Bench2Drive~\cite{jia2024bench2drive}                & {42.35}           & {15}                  & {157.94}       & {46.01}         \\
                            & DriveCombo                   & 26.68           & 13.33                & 116.53        & 33.96       \\
\bottomrule                            
\end{tabular}
\end{table*}

\subsection{Potential of Closed-loop Planning Evaluation}
In the main paper, we show that the DriveCombo dataset can evaluate the reasoning of traffic rules in MLLMs through a question and answer format. We further note that the dataset is also suitable for closed-loop trajectory evaluation of end to end (E2E) models, making it a comprehensive benchmark that covers both reasoning and planning capabilities. 
DriveCombo’s scenarios provide highly structured descriptions of traffic events, enabling plug-and-play integration with the Bench2Drive framework~\cite{jia2024bench2drive}. As shown in Figure~\ref{fig: supp_close_loop}, we can seamlessly convert DriveCombo’s scenarios into closed-loop test cases to examine E2E models under real rule constraints. Specifically, we first align the original OpenScenario scenes to the Bench2Drive format and import them into CARLA. We then use the PDM-lite expert model~\cite{sima2023drivelm} to generate ground-truth trajectories and evaluate various E2E models.

Using the Bench2Drive evaluation setup, we test E2E methods on fifteen short routes. All metrics follow the Bench2Drive evaluation standard. As shown in Table~\ref{tab:supp_eval_on_bench2drive}, both E2E models exhibit substantial performance drops on DriveCombo. For example, the Driving Score of UniAD Base~\cite{hu2023uniad} decreases from 45.81 to 27.59 and VAD~\cite{Jiang2023VADVS} decreases from 42.35 to 26.68. Metrics related to efficiency and comfort also decline significantly. These results indicate that DriveCombo presents greater challenges in terms of rule complexity, traffic interaction and dynamic risk, and imposes stricter requirements on E2E methods. We plan to further expand the DriveCombo dataset to enable more comprehensive and systematic evaluation of trajectory planning capabilities.

\subsection{Performance of MLLMs across countries on DriveCombo.}
\label{sec:supp_multi_country_results}
Results reported in the main paper are averaged over five countries to provide an overall assessment, while per-country performance is reported in Table~\ref{tab:country_performance} to reveal cross-country performance variations. The table reports average performance over 8 open-source models and 6 proprietary models evaluated in the main paper, highlighting performance variations across different national traffic rule systems. 
Overall, models achieve relatively higher accuracy on the USA and China subsets, while performance is comparatively lower on Japan and Australia, indicating potential challenges posed by country-specific regulatory structures and rule distributions.

\subsection{Multi-Rule Compositional Reasoning}
Table~\ref{tab:comb3_comb4_comb5_result} evaluates the performance of MLLMs in multi-rule driving scenarios. As shown, under the \#Rules = 3 setting, both open-source and proprietary models achieve relatively high scores. For example, larger open-source models like GLM-4.5V~\cite{vteam2025glm45v} reach about 68.15 accuracy (acc.) on L2–L4, GPT-5 Pro~\cite{openai2025gpt5} further improves to 71.09.
However, when the number of rules increases to \#Rules = 4, all models show noticeable degradation, particularly on the challenging L4 and L5 tasks. In the high-complexity setting of \#Rules = 5, even the best proprietary model, GPT-5 Pro, achieves only 34.78 acc in L5; the best open-source model, GLM-4.5V, reaches 32.78, also showing a clear performance degradation.
Overall, as the complexity of the scenario increases (i.e., \#Rules from 3 to 5), all models exhibit performance declines, indicating that DriveCombo introduces substantial difficulty in traffic-rule understanding and cross-entity interaction reasoning, and that current MLLMs still struggle with high-complexity driving scenarios.

\begin{table*}[h]
\centering
\vspace{-3mm}
\caption{\textbf{Performance of MLLMs across countries on DriveCombo.} We report average results of 8 open-source and 6 proprietary models used in the main paper.}
\vspace{-3mm}
\label{tab:country_performance}
\setlength{\tabcolsep}{8.5pt}
\begin{tabular}{lccccc|ccccc}
\toprule
& \multicolumn{5}{c}{\textbf{Open-source Models (Avg.)}} 
& \multicolumn{5}{c}{\textbf{Proprietary Models (Avg.)}} \\
\cmidrule(lr){2-6} \cmidrule(lr){7-11}
\textbf{Country} 
& \textbf{L1} & \textbf{L2} & \textbf{L3} & \textbf{L4} & \textbf{L5}
& \textbf{L1} & \textbf{L2} & \textbf{L3} & \textbf{L4} & \textbf{L5} \\
\midrule
USA        & 71.36 & 67.69 & 62.26 & 58.98 & 35.11 & 86.29 & 79.50 & 72.93 & 68.69 & 42.71 \\
China      & 72.38 & 67.25 & 63.01 & 59.99 & 34.76 & 85.89 & 79.22 & 73.03 & 68.95 & 42.41 \\
UK         & 70.07 & 66.56 & 62.48 & 56.33 & 33.58 & 83.55 & 78.80 & 72.66 & 67.91 & 41.29 \\
Japan      & 63.60 & 58.06 & 55.77 & 52.28 & 30.11 & 73.86 & 71.56 & 65.36 & 63.03 & 37.26 \\
Australia  & 61.75 & 55.18 & 51.29 & 49.96 & 29.68 & 79.05 & 72.09 & 64.03 & 62.39 & 37.90 \\
\midrule
Average    & 67.83 & 62.95 & 58.96 & 55.51 & 32.65 & 81.73 & 76.23 & 69.60 & 66.20 & 40.31 \\
\bottomrule
\end{tabular}
\end{table*}
\vspace{-3mm}

\setlength{\tabcolsep}{4.5pt}
\begin{table*}
\centering
\caption{\textbf{Performance of MLLM in Complex Driving Scenarios with  Different Numbers of Rules.} ``\#Rules" denotes number of traffic rules in each scenario. \colorbox[HTML]{ADD88D}{Green} and \colorbox[HTML]{E3F2D9}{light green} mark the best and second-best open-source models, while \colorbox[HTML]{B5C6EA}{blue} and \colorbox[HTML]{D9E1F4}{light blue} indicate the best and second-best proprietary models.}
\label{tab:comb3_comb4_comb5_result}
\begin{tabular}{lc|cccc|cccc|cccc}
\toprule
\multirow{2}{*}{\textbf{Model}} & \multicolumn{1}{l|}{\multirow{2}{*}{\textbf{Size}}} & \textbf{L2} & \textbf{L3} & \textbf{L4} & \textbf{L5} & \textbf{L2} & \textbf{L3} & \textbf{L4} & \textbf{L5} & \textbf{L2} & \textbf{L3} & \textbf{L4} & \textbf{L5} \\
 & \multicolumn{1}{l|}{} & \multicolumn{4}{c|}{\textbf{\#Rules=3}} & \multicolumn{4}{c|}{\textbf{\#Rules=4}} & \multicolumn{4}{c}{\textbf{\#Rules=5}} \\
\midrule
\multicolumn{14}{c}{\textit{Open-source Models}}\\
Gemma 3 & 4B & 47.11 & 44.57 & 41.89 & 20.05 & 44.40 & 42.52 & 39.91 & 16.57 & 40.93 & 38.12 & 32.06 & 13.89 \\
Gemma 3 & \multicolumn{1}{l}{12B} & 56.30 & 54.85 & 50.14 & 24.41 & 45.97 & 43.82 & 43.31 & 23.43 & 43.41 & 41.57 & 37.72 & 20.45 \\
Gemma 3 & 27B & 60.74 & 58.18 & 54.23 & 33.05 & 52.56 & 51.96 & 47.68 & 31.33 & 48.59 & 45.86 & 42.07 & 27.51 \\
Llama 3.2 & 11B & 48.38 & 43.46 & 38.29 & 17.57 & 38.55 & 35.21 & 34.68 & 13.29 & 34.59 & 25.96 & 22.87 & 8.71 \\
Qwen3-VL & 2B & 51.96 & 50.38 & 48.61 & 24.12 & 44.82 & 41.41 & 36.13 & 22.60 & 42.39 & 40.15 & 34.48 & 18.92 \\
Qwen3-VL & 8B & 62.84 & 59.90 & 54.54 & 31.81 & 55.87 & 55.09 & 52.20 & 30.12 & 50.91 & 46.16 & 38.76 & 24.67 \\
Qwen3-VL & 32B & \cellcolor[HTML]{E3F2D9}69.87 & \cellcolor[HTML]{E3F2D9}65.95 & \cellcolor[HTML]{E3F2D9}60.43 & \cellcolor[HTML]{E3F2D9}35.34 & \cellcolor[HTML]{E3F2D9}65.31 & \cellcolor[HTML]{E3F2D9}60.64 & \cellcolor[HTML]{E3F2D9}54.60 & \cellcolor[HTML]{E3F2D9}32.91 & \cellcolor[HTML]{E3F2D9}55.78 & \cellcolor[HTML]{E3F2D9}54.95 & \cellcolor[HTML]{E3F2D9}49.97 & \cellcolor[HTML]{E3F2D9}31.38 \\
GLM-4.5V & 106B & \cellcolor[HTML]{ADD88D}72.67 & \cellcolor[HTML]{ADD88D}66.85 & \cellcolor[HTML]{ADD88D}64.93 & \cellcolor[HTML]{ADD88D}40.64 & \cellcolor[HTML]{ADD88D}66.71 & \cellcolor[HTML]{ADD88D}62.24 & \cellcolor[HTML]{ADD88D}56.40 & \cellcolor[HTML]{ADD88D}36.71 & \cellcolor[HTML]{ADD88D}58.48 & \cellcolor[HTML]{ADD88D}56.95 & \cellcolor[HTML]{ADD88D}51.07 & \cellcolor[HTML]{ADD88D}32.78 \\
\midrule
\multicolumn{14}{c}{\textit{Proprietary Models}} \\
Gemini 2.5 flash & - & 71.12 & 67.48 & 61.42 & 36.07 & 62.57 & 54.47 & 52.86 & 35.20 & 56.70 & 48.45 & 46.18 & 30.80 \\
Gemini 2.5 pro & - & 75.99 & \cellcolor[HTML]{D9E1F4}69.72 & 65.10 & \cellcolor[HTML]{D9E1F4}41.26 & 65.24 & 61.35 & 55.91 & 36.54 & \cellcolor[HTML]{D9E1F4}60.22 & 56.02 & \cellcolor[HTML]{D9E1F4}54.87 & 33.82 \\
Claude sonnet 4.5 & - & \cellcolor[HTML]{D9E1F4}77.40 & 68.23 & \cellcolor[HTML]{D9E1F4}65.41 & 41.19 & \cellcolor[HTML]{D9E1F4}65.97 & \cellcolor[HTML]{D9E1F4}63.73 & \cellcolor[HTML]{D9E1F4}56.15 & \cellcolor[HTML]{D9E1F4}36.75 & 59.19 & \cellcolor[HTML]{D9E1F4}57.25 & 53.60 & \cellcolor[HTML]{D9E1F4}34.28 \\
GPT-5 nano & - & 58.54 & 57.68 & 55.92 & 35.70 & 50.42 & 50.38 & 46.59 & 30.97 & 49.70 & 47.56 & 45.78 & 28.04 \\
GPT-5 mini & - & 61.37 & 68.46 & 60.43 & 37.62 & 61.35 & 53.14 & 51.07 & 32.81 & 47.74 & 46.57 & 45.01 & 32.23 \\
GPT-5 pro & - & \cellcolor[HTML]{B5C6EA}77.93 & \cellcolor[HTML]{B5C6EA}69.83 & \cellcolor[HTML]{B5C6EA}65.71 & \cellcolor[HTML]{B5C6EA}42.19 & \cellcolor[HTML]{B5C6EA}68.67 & \cellcolor[HTML]{B5C6EA}64.93 & \cellcolor[HTML]{B5C6EA}58.25 & \cellcolor[HTML]{B5C6EA}37.25 & \cellcolor[HTML]{B5C6EA}61.29 & \cellcolor[HTML]{B5C6EA}59.15 & \cellcolor[HTML]{B5C6EA}55.00 & \cellcolor[HTML]{B5C6EA}34.78\\
\bottomrule
\end{tabular}
\end{table*}

\begin{table*}
\centering
\footnotesize
\caption{\textbf{Prompt of MCQ Generation of Level 1.}}
\label{tab:prompt_of_mcq_generation_1}
\begin{tabular}{p{0.95\linewidth}}
\toprule
\textbf{System Persona} \\ \midrule
    You are a question-generation assistant proficient in traffic regulations.
    
    I will give you a driving rule. Please generate an autonomous driving test case based on the given traffic regulation clause. Output the result in JSON format, including the following fields: {scenario description, question stem, options (one correct option and three distractors), question logic, correct answer option, and an explanation of the correct answer}. The question design should follow these principles:

    1. Ensure that the options have similar semantics to prevent the model from guessing based on surface-level wording.

    2. To increase the difficulty of the question, avoid including any explicit safety-related cues or contextual hints in the scenario description, question stem, or options that could reveal the correct answer.

    3. Use vague or implicit semantic descriptions: remove direct hints and instead imply trigger conditions through sensor data or subtle scene details to stimulate the model's reasoning and judgment abilities.

    4. When designing the correct answer, create options that are “counterintuitive yet reasonable.” Avoid making “the safest choice” automatically the correct one — the goal is to make “safest” not equal to “most compliant.”

    5. If a regulation contains multiple triggering conditions, select only one to construct the test scenario. For example, if a regulation states: “When driving at night, in rain or fog, or when visibility is poor and the safe distance cannot be confirmed, lane changes are prohibited,” choose only one weather condition (e.g., rain) as the trigger.

    Please remember that you should output json without any other output, the format of json is:
    
    \quad\{
    
           \qquad ``Scenario Description": ``...",
        
           \qquad ``Question Stem": ``...",
        
           \qquad ``Options": \{
        
    \qquad\qquad``A": ``...",
            
    \qquad\qquad``B": ``...",
            
    \qquad\qquad``C": ``...",
            
    \qquad\qquad``D": ``..."
            
            \qquad\qquad\},
        
           \qquad ``Question Design Logic": ``...",
        
           \qquad ``Correct Answer Option": ``...",
        
           \qquad ``Explanation of the Correct Answer": ``..."
           
    \quad\}.

    \textit{Example 1} :
    
    Input Rule: Other vehicles and pedestrians shall yield to vehicles performing emergency duties.

    Output JSON: 
    
    \quad\{
    
  \qquad``Scenario Description": ``An autonomous vehicle is driving on an urban main road. The ambulance is rapidly approaching from behind. The current lane is congested, and surrounding vehicles are reacting differently—some slow down and move aside, while others maintain their speed.",
  
  \qquad``Question Stem": ``As the driver, how should you respond correctly in this situation?",
  
  \qquad``Options": \{
  
    \qquad\qquad``A": ``Immediately pull over and stop to make way for the ambulance to pass first.",
    
    \qquad\qquad``B": ``Maintain the current speed and avoid changing lanes to prevent traffic confusion.",
    
    \qquad\qquad``C": ``Accelerate to leave the area quickly and avoid crossing paths with the ambulance.",
    
    \qquad\qquad``D": ``Continue driving along the planned route and only consider yielding when the ambulance is within 5 meters."
    
  \qquad\qquad \},
  
  \qquad``Question Design Logic": ``This question examines whether an autonomous driving system adheres to legal requirements upon detecting an emergency vehicle, rather than merely following standard traffic flow strategies. The scenario implies the presence of an “emergency-response vehicle” through auditory and visual cues—such as sirens and red-blue flashing lights—without explicitly stating instructions like “pull over immediately.” This design tests whether the system can infer the obligation to yield based on contextual features.Option A represents lawful behavior in accordance with Article 53 of the Road Traffic Safety Law. Options B and D, though appearing to “maintain order,” fail to promptly fulfill the duty to yield. Option C reflects a common “counterintuitive yet unlawful” response—accelerating to avoid interference, which in fact obstructs emergency passage.The key focus of the question lies in whether the model can correctly prioritize between “traffic flow stability” and “emergency right-of-way,” recognizing the higher legal precedence of the latter.",
  
  \qquad``Correct Answer Option": ``B",
  
  \qquad``Explanation of the Correct Answer": ``According to Article 53 of the Road Traffic Safety Law, other vehicles and pedestrians must yield to vehicles performing emergency duties. When an autonomous system detects emergency vehicles such as ambulances, fire trucks, or police cars, it should proactively slow down and pull over to allow them to pass. Option A correctly reflects the yielding principle. Option B avoids lane changes but fails to fulfill the yielding obligation; Option C creates a potential safety hazard by accelerating; Option D's conditional yielding violates the requirement for immediate cooperation."
  
\quad\}

\textit{Example 2} : ... 

\\
\midrule 
\textbf{User Input} \\ \midrule
    Input Rule: \{rule\_content\}
    
    Output JSON:
\\
\bottomrule
\end{tabular}
\end{table*}

\begin{table*}
\centering
\footnotesize
\caption{\textbf{Prompt of MCQ Generation of Level 2-4.}}
\label{tab:prompt_of_mcq_generation_2_3_4}
\begin{tabular}{p{0.95\linewidth}}
\toprule
\textbf{System Persona} \\ \midrule
    You are a question-generation assistant proficient in traffic regulations.
    
    I will give you \textit{\{two / three / four / five\}} driving rules. Please generate an autonomous driving test case based on the \textit{\{two / three / four / five\}} given traffic regulation clauses. Output the result in JSON format, including the following fields: {scenario description, question stem, options (one correct option and three distractors), question logic, correct answer option, and an explanation of the correct answer}. The question design should follow these principles:

    1. Ensure that the options have similar semantics to prevent the model from guessing based on surface-level wording.

    2. To increase the difficulty of the question, avoid including any explicit safety-related cues or contextual hints in the scenario description, question stem, or options that could reveal the correct answer.

    3. Use vague or implicit semantic descriptions: remove direct hints and instead imply trigger conditions through sensor data or subtle scene details to stimulate the model's reasoning and judgment abilities.

    4. When designing the correct answer, create options that are “counterintuitive yet reasonable.” Avoid making “the safest choice” automatically the correct one — the goal is to make “safest” not equal to “most compliant.”

    5. If a regulation contains multiple triggering conditions, select only one to construct the test scenario. For example, if a regulation states: “When driving at night, in rain or fog, or when visibility is poor and the safe distance cannot be confirmed, lane changes are prohibited,” choose only one weather condition (e.g., rain) as the trigger.

    Please remember that you should output json without any other output, the format of json is:
    
    \quad\{
    
           \qquad ``Scenario Description": ``...",
        
           \qquad ``Question Stem": ``...",
        
           \qquad ``Options": \{
        
    \qquad\qquad``A": ``...",
            
    \qquad\qquad``B": ``...",
            
    \qquad\qquad``C": ``...",
            
    \qquad\qquad``D": ``..."
            
            \qquad\qquad\},
        
           \qquad ``Question Design Logic": ``...",
        
           \qquad ``Correct Answer Option": ``...",
        
           \qquad ``Explanation of the Correct Answer": ``..."
           
    \quad\}.

    \textit{Example 1} :
    
    Input Rule 1: When driving on a narrow bridge ...
    
    Input Rule 2: When lane markings are dashed ... 
    
    Input Rule 3 (if necessary) : ...

    Input Rule 4 (if necessary) : ...
    
    Input Rule 5 (if necessary) : ...

    Output JSON: 
    
    \quad\{
    
  \qquad``Scenario Description": ``The ego vehicle is driving along a two-way ...",
  
  \qquad``Question Stem": ``As the driver, how should you respond correctly in this situation?",
  
  \qquad``Options": \{
  
    \qquad\qquad``A": ``Quickly overtake the front ...",
    
    \qquad\qquad``B": ``Slow down, maintain a safe ...",
    
    \qquad\qquad``C": ``Briefly use the oncoming ...",
    
    \qquad\qquad``D": ``Honk in advance to alert ..."
    
  \qquad\qquad \},
  
  \qquad``Question Design Logic": ``This question combines two traffic ...",
  
  \qquad``Correct Answer Option": ``B",
  
  \qquad``Explanation of the Correct Answer": ``Option B represents compliant behavior: maintaining ..."
  
\quad\}

\textit{Example 2} : ... 

\\
\midrule 
\textbf{User Input} \\ \midrule
    Input Rule 1: \{rule\_content\_1\}
    
    Input Rule 2: \{rule\_content\_2\}
    
    Input Rule 3 (if necessary): \{rule\_content\_3\}
    
    Input Rule 4 (if necessary): \{rule\_content\_4\}
    
    Input Rule 5 (if necessary): \{rule\_content\_5\}
    
    Output JSON:
\\
\bottomrule
\end{tabular}
\end{table*}

\begin{table*}
\centering
\footnotesize
\caption{\textbf{Prompt of MCQ Generation of Level 5.}}
\label{tab:prompt_of_mcq_generation_5}
\begin{tabular}{p{0.9\linewidth}}
\toprule
\textbf{System Persona} \\ \midrule
 I will give you \textit{\{two / three / four / five\}} driving rules. Please generate an autonomous driving test case based on the \textit{\{two / three / four / five\}} given traffic regulation clauses. The case should create a priority conflict that requires the test taker to make a judgment. Output the result in JSON format, including the following fields: {scenario description, question stem, options (one correct option and three distractors), question logic, correct answer option, and an explanation of the correct answer}. The question design should follow these principles:

    1. You must select the regulation with the higher priority from the \textit{\{two / three / four / five\}} given inputs as the correct answer when creating the question. The priority hierarchy of traffic regulations is as follows:
    Pedestrian life safety $>$ Emergency avoidance vehicles $>$ On-site command personnel $>$ Traffic lights $>$ Traffic signs $>$ Road markings $>$ Interactive right-of-way $>$ Defensive driving $>$ Emergency exceptions.

    2. Ensure that the answer options have similar semantic meanings, to prevent the model from guessing based on surface-level wording.

    3. To increase the difficulty level, avoid explicit safety cues in the scenario description, question stem, and answer options. Do not include context that clearly suggests which choice is safer or more correct.

    4. The scenario should not fully satisfy any single rule; instead, place it in a borderline condition so that the test taker must decide whether it “qualifies.” For example, when describing environmental conditions, be ambiguous—avoid direct statements like “poor visibility” or “low visibility.” Instead, describe how sensors behave, letting the test taker infer whether the rule should be triggered.

    5. When designing the correct answer, make it “counterintuitive but reasonable.” Avoid making “the safest option” always the correct one. The correct answer should reflect regulatory compliance, not simply maximum safety.

    6. If a regulation contains multiple trigger conditions, choose only one to build the test scenario. For example, if a rule states: “When driving at night, in rain/fog, or under poor visibility where safe distance cannot be confirmed, lane changes are prohibited,” select just one condition (e.g., nighttime) as the trigger.

    7. The options must represent specific driving actions. If the \textit{\{two / three / four / five\}} input regulations concern lane-change behavior, the scenario should feature a multi-lane road, and the answer options could be: changing lanes to the left, changing lanes to the right, keeping the current lane, or other relevant maneuvers.

    Please remember that you should output json without any other output, the format of json is:
    
    \quad\{
    
           \qquad ``Scenario Description": ``...",
        
           \qquad ``Question Stem": ``...",
        
           \qquad ``Options": \{
        
    \qquad\qquad``A": ``...",
            
    \qquad\qquad``B": ``...",
            
    \qquad\qquad``C": ``...",
            
    \qquad\qquad``D": ``..."
            
            \qquad\qquad\},
        
           \qquad ``Question Design Logic": ``...",
        
           \qquad ``Correct Answer Option": ``...",
        
           \qquad ``Explanation of the Correct Answer": ``..."
           
    \quad\}.

    \textit{Example 1} :
    
    Input Rule 1: When it is nighttime ...
    
    Input Rule 2: When the driver needs ... 
    
    Input Rule 3 (if necessary) : ...

    Input Rule 4 (if necessary) : ...
    
    Input Rule 5 (if necessary) : ...

    Output JSON: 
    
    \quad\{
    
  \qquad``Scenario Description": ``The ego vehicle is ...",
  
  \qquad``Question Stem": ``As the driver, how should you respond correctly in this situation?",
  
  \qquad``Options": \{
  
    \qquad\qquad``A": ``Take advantage of a brief gap ...",
    
    \qquad\qquad``B": ``Stay in the current lane ...",
    
    \qquad\qquad``C": ``Slightly adjust the driving ...",
    
    \qquad\qquad``D": ``Brake firmly and stop ..."
    
  \qquad\qquad \},
  
  \qquad``Question Design Logic": ``Regulation 1: 'When it is nighttime, raining, foggy, or when the safe distance cannot ...",
  
  \qquad``Correct Answer Option": ``B",
  
  \qquad``Explanation of the Correct Answer": ``Although the fallen cone ahead is an obstacle ..."
  
\quad\}

    \textit{Example 2} : ...

\\
\midrule 
\textbf{User Input} \\ \midrule
    Input Rule 1: \{rule\_content\_1\}
    
    Input Rule 2: \{rule\_content\_2\}
    
    Input Rule 3 (if necessary): \{rule\_content\_3\}
    
    Input Rule 4 (if necessary): \{rule\_content\_4\}
    
    Input Rule 5 (if necessary): \{rule\_content\_5\}
    
    Output JSON:
\\
\bottomrule
\end{tabular}
\end{table*}

\begin{table*}
\centering
\footnotesize
\caption{\textbf{Prompt of MCQ Generation Check.}}
\label{tab:prompt_of_mcq_generation_check}
\begin{tabular}{p{0.95\linewidth}}
\toprule
\textbf{System Persona} \\ \midrule
  Your task is to act as a rigorous evaluator of a generated driving-rule MCQ question. I will provide one question in JSON format. Your goal is to determine whether the question is logically valid and fully consistent with the traffic rules, based on four criteria:
    
    1. Correctness of the Answer:
    
       - The “Correct Answer Option” must be logically correct.
       
       - The explanation must be sound, and no other option may also be correct.
    
    2. Faithfulness to the Input Rules:
    
       - The scenario must accurately incorporate all and only the rules implied in the question design.
         
       - No part of the scenario should contradict the rule conditions.
    
    3. Quality of the Options:
    
       - There must be exactly four options (A, B, C, D).
       
       - The distractors must be plausible but incorrect.
       
       - No option may be ambiguous or require unstated assumptions.
    
    4. Structural Completeness:
    
       - The JSON must include: Scenario Description, Question Stem, Options, 
         Correct Answer Option, and Explanation.
         
       - All parts must be internally coherent.
    
    After checking all criteria:
    
    - If ANY criterion fails → Output 0, followed by the reasoning.
    
    - If ALL criteria pass → Output 1, followed by the reasoning.

    \textit{Example 1} :

    Input Question Json: 
    
    \quad\{
    
        \qquad "Scenario Description": "A foggy day with visibility below 40 meters.",
        
        \qquad "Question Stem": "What is the correct speed?",
        
        \qquad "Options": \{
        
        \qquad\qquad "A": "70 km/h",
        
        \qquad\qquad "B": "30 km/h",
        
        \qquad\qquad "C": "Turn right",
        
        \qquad\qquad "D": "High beam"
        
        \qquad \},
        
        \qquad "Question Design Logic": "Low-visibility speed rule.",
        
        \qquad "Correct Answer Option": "A",
        
        \qquad "Explanation of the Correct Answer": "High speed is acceptable in fog."
        
    \quad\}

    Output Decision: 0
    
    Output Reasoning: The selected answer violates low-visibility speed rules; distractors include irrelevant actions; the explanation contradicts the rule. Therefore, the question is invalid.

    \textit{Example 2} : ...

\\
\midrule 
\textbf{User Input} \\ \midrule
    Input Question Json: \{ question json\}
    
    Output Decision:
    
    Output Reasoning:
\\
\bottomrule
\end{tabular}
\end{table*}

\begin{table*}
\centering
\footnotesize
\caption{\textbf{Prompt of Semantic Structuring.}}
\label{tab:prompt-semantic-structuring}
\begin{tabular}{p{0.95\linewidth}}
\toprule
\textbf{System Persona} \\ \midrule
You are an expert assistant proficient in analyzing and structuring traffic regulations.

I will give you a natural-language traffic rule. Your task is to transform this rule into a normalized atomic structure by extracting exactly the following five fields:

1. \textbf{Rule Content}:  

   A concise restatement of the core meaning of the rule without ambiguity.  
   The restatement must follow the template: ``When [condition], the driver may / must / must not [action].''

2. Perceptual Type:  

   Categorize the rule as  
   
   \quad ``static''  (triggered by signs, markings, road type, fixed conditions), or  
   
   \quad ``dynamic'' (triggered by interactions with other vehicles, pedestrians, or changing traffic flow).

3. Norm Type:  

   Determine whether the rule is  
   
   \quad ``obligatory''  (the driver must perform the action),  
   
   \quad ``forbidden''   (the driver must not perform the action), or  
   
   \quad ``permissive''  (the driver may perform the action).

4. Action Type:  

   The primary driving action described by the rule (e.g., overtake, left turn, right turn, lane change, merge, acceleration, deceleration, signaling, yielding, parking, etc.).

5. Numeric Constraints (if applicable):  

   Extract any explicit speed limits.

You must output the result strictly in the following JSON format:

\quad\{

\qquad ``rule\_content": ``...",  

\qquad ``perceptual\_type": ``...",  

\qquad ``norm\_type": ``...",  

\qquad ``action\_type": ``...",  

\qquad ``numeric\_constraints": ``..."  

\quad\}

Guidelines:  

- Do NOT add any additional fields.  

- Do NOT invent content beyond what the rule states.  

- Keep each field concise but semantically complete.

\textit{Example 1} :

Input Rule: When visibility is below 50 meters due to heavy fog, the driver must reduce speed to at most 30 km/h.

Output JSON:

\quad\{

\qquad ``rule\_content": ``When visibility falls below 50 meters, must slow to 30 km/h. ",

\qquad ``perceptual\_type": ``static",

\qquad ``norm\_type": ``obligatory",

\qquad ``action\_type": ``speed limit",

\qquad ``numeric\_constraints": \{

\qquad \qquad ``upper bound": 30

\qquad \qquad ``lower bound": 0

  \qquad\qquad \},

\quad\}

\textit{Example 2} :

Input Rule: Drivers must not overtake when a solid yellow centerline is present.

Output JSON:

\quad\{

\qquad ``rule\_content": ``When a solid yellow centerline is present, the driver must not overtake.",

\qquad ``perceptual\_type": ``static",

\qquad ``norm\_type": ``forbidden",

\qquad ``action\_type": ``overtake",

\qquad ``numeric\_constraints": ``none"

\quad\}

\\
\midrule
\textbf{User Input} \\ \midrule
Input Rule: \{traffic\_rule\}

Output JSON:
\\
\bottomrule
\end{tabular}
\end{table*}

\begin{table*}
\centering
\footnotesize
\caption{\textbf{Prompt of Rule Coexistence Validation.}}
\label{tab:prompt_combination_check}
\begin{tabular}{p{0.95\linewidth}}
\toprule
\textbf{System Persona} \\ \midrule

    I aim to combine atomic traffic regulations to form new, more complex composite regulations. I will input \textit{\{two / three / four / five\}} such atomic regulations, and your task is to determine whether these \textit{\{two / three / four / five\}} input regulations are compatible in terms of both scenario and strategy—that is, whether they can coexist within the same context. The specific detection procedure is as follows:

    1. For each traffic regulation, extract the context refers to the situational conditions under which the regulation applies.

    2. Examine whether their contexts are mutually exclusive. Two contexts are considered mutually exclusive if the physical scenarios they describe cannot coexist in the same space-time.
  
    After the above checks, if the contexts of these \textit{\{two / three / four / five\}} regulations are mutually exclusive, output 0; if they are compatible, output 1, along with the reasoning.

\textit{Example 1} :

Input Rule 1: ``In foggy conditions, motor vehicles shall reduce their driving speed.''  

Input Rule 2: ``When the road surface is icy, motor vehicles shall reduce their driving speed.''

Output: 1.  
Reasoning: Fog and icy road conditions may occur simultaneously; both describe adverse weather conditions and can coexist.

\textit{Example 2} :

Input Rule 1: ``When an on-ramp or branch road has an acceleration lane and traffic signs permit, the ego vehicle may merge into the main road.''  

Input Rule 2: ``When there is a “No Entry” or “No Merging” sign, the ego vehicle is prohibited from merging into the main road.''

Input Rule 3: ``When the area between the ramp or service road and the main road is marked with a solid line or a hatched/guide line area, the ego vehicle is prohibited from merging into the main road.''

Output: 0.  
Reasoning: These rules cannot be combined because they describe mutually exclusive scenarios. Rule 1 only applies when merging is permitted (acceleration lane present and signs allow), while Rule 2 prohibits merging when “No Entry” or “No Merging” signs are present, and Rule 3 prohibits merging when solid or hatched lines are present. The “permitted” condition of Rule 1 and the “prohibited” conditions of Rules 2 and 3 cannot exist simultaneously in the same place and time (permission and prohibition cannot coexist). Although Rules 2 and 3 can coexist and jointly reinforce the prohibition, Rule 1 conflicts with both, so overall, they cannot logically coexist or be combined.

\textit{Example 3} : ...

\\
\midrule
\textbf{User Input} \\ \midrule
Input Rule 1: \{rule\_1\} \\
Input Rule 2: \{rule\_2\} \\
Input Rule 3 (if necessary): \{rule\_3\} \\
Input Rule 4 (if necessary): \{rule\_4\} \\
Input Rule 5 (if necessary): \{rule\_5\} \\
Output:
\\
\bottomrule
\end{tabular}
\end{table*}

\begin{table*}
\centering
\footnotesize
\caption{\textbf{Prompt of DSL Translation.}}
\label{tab:prompt-dsl-translation}
\begin{tabular}{p{0.95\linewidth}}
\toprule
\textbf{System Persona} \\ \midrule
You are an expert assistant in autonomous driving test scenario generation.

You will be given:  

(1) a textual scenario description derived from traffic rules;  

(2) a DSL specification that defines the syntax and structure of executable driving scenarios;  

(3) an example DSL illustrating how rule conditions map to YAML-based scene definitions.

Your task is to translate the scenario description into a structured semantic DSL representation.  

The generated DSL must explicitly encode:

1. Entities: ego vehicle, other vehicles, pedestrians, static obstacles, traffic lights, etc.  

2. Spatial Relations: relative positions (ahead of, behind, left of, right of), lane index,  
   distance relations, orientation, and spatial layout.  

3. Environment: road type, lane geometry, weather, visibility, time of day.  

4. Behavioral Trajectories: initial poses, intended paths, speed profiles, and motion behaviors.

5. Rule Satisfaction: every constraint implied by the traffic rules must appear explicitly in the DSL.

Final Requirement:  
Output \emph{only} the YAML DSL scenario.  
Do \emph{not} output explanations, comments, or any additional text.

\textit{Example 1} :

\textbf{Input Scenario Description:} 
``The ego vehicle is traveling in the leftmost lane of a three-lane urban road. The navigation system indicates that the vehicle needs to exit at the next intersection 200 meters ahead, which requires the vehicle to be in the rightmost lane. The vehicles in the adjacent lanes are moving at similar speeds.''

\textbf{Input Traffic Rules:} Navigation requires lane change toward the rightmost lane before reaching an exit.

\textbf{Output DSL (YAML):}

\quad\texttt{environment:} \\
\qquad\texttt{weather: sunny} \\
\qquad\texttt{time: daytime} \\

\quad\texttt{road\_network:} \\
\qquad\texttt{road\_type: intersection} \\
\qquad\texttt{road\_marker: solid\_line} \\
\qquad\texttt{traffic\_signs:} \\
\qquad\qquad\texttt{- traffic\_light} \\

\quad\texttt{actors:} \\

\qquad\texttt{- id: ego} \\
\qquad\qquad\texttt{type: car} \\
\qquad\qquad\texttt{position:} \\
\qquad\qquad\qquad\texttt{reference: intersection} \\
\qquad\qquad\qquad\texttt{relation: behind} \\
\qquad\qquad\texttt{behavior: turn\_right} \\

\qquad\texttt{- id: vehicle\_1} \\
\qquad\qquad\texttt{type: car} \\
\qquad\qquad\texttt{position:} \\
\qquad\qquad\qquad\texttt{reference: intersection} \\
\qquad\qquad\qquad\texttt{relation: front} \\
\qquad\qquad\texttt{behavior: go\_forward} \\

\qquad\texttt{- id: vehicle\_2} \\
\qquad\qquad\texttt{type: car} \\
\qquad\qquad\texttt{position:} \\
\qquad\qquad\qquad\texttt{reference: ego} \\
\qquad\qquad\qquad\texttt{relation: right} \\
\qquad\qquad\texttt{behavior: turn\_right} \\

\qquad\qquad\texttt{...} \\

\quad\texttt{oracle:} \\
\qquad\texttt{longitudinal: go\_forward} \\
\qquad\texttt{lateral: keep\_lane} \\

\textit{Example 2} :...
\\
\midrule
\textbf{User Input} \\ \midrule
Scenario Description: \{scene\_text\} \\
Traffic Rules: \{rule\_text\} \\
Output YAML DSL:
\\
\bottomrule
\end{tabular}
\end{table*}

\begin{table*}
\centering
\footnotesize
\caption{\textbf{Quality Scoring Prompt for Semantic Structuring.}}
\label{tab:prompt-quality-scoring-semantic}
\begin{tabular}{p{0.95\linewidth}}
\toprule
\textbf{System Persona} \\ \midrule
You are an expert evaluator in traffic-rule parsing.  
You will be given (1) the exact prompt used to generate the semantic-structuring output,  
and (2) the model output produced by that prompt.

Your task is to assign a quality score in the range \texttt{[0, 1]} evaluating:

1. \textbf{Correctness}: whether the structured fields reflect the original rule described in the prompt.  

2. \textbf{Completeness}: whether all required fields (rule content, perceptual type, norm type, action type, numeric constraints) are present. 

3. \textbf{Fidelity}: whether the model output introduces no hallucinations or distortions relative to the prompt.

Output \textbf{only a floating-point score} in \texttt{[0,1]}. 

Do \textbf{not} output explanations or text.  

The final output format must be:

\quad\texttt{<score>}
\\
\midrule
\textbf{User Input} \\ \midrule
Semantic Structuring Prompt \\
Semantic Structuring Output \\
Output Score:

\\
\bottomrule
\end{tabular}
\end{table*}

\begin{table*}
\centering
\footnotesize
\caption{\textbf{Quality Scoring Prompt for Coexistence Validation.}}
\label{tab:prompt-quality-scoring-coexistence}
\begin{tabular}{p{0.95\linewidth}}
\toprule
\textbf{System Persona} \\ \midrule
You are an expert evaluator in multi-rule compatibility reasoning.  
You will be given (1) the exact prompt used to perform coexistence validation,  
and (2) the model output generated by that prompt.

Your task is to assign a score in \texttt{[0,1]} evaluating:

1. \textbf{Logical validity}: whether the compatibility judgment (0/1) matches the true feasibility implied by the rules.  

2. \textbf{Scenario correctness}: whether the reasoning in the output aligns with the real-world spatial and temporal constraints described in the prompt.  

3. \textbf{Fidelity to rules}: whether the model correctly interprets the rules without inventing new conditions.

Output \textbf{only a score} in \texttt{[0,1]}.  
No text, no reasoning.

\quad\texttt{<score>}
\\
\midrule
\textbf{User Input} \\ \midrule
Coexistence Validation Prompt \\
Coexistence Validation Output \\
Output Score:

\\
\bottomrule
\end{tabular}
\end{table*}

\begin{table*}
\centering
\footnotesize
\caption{\textbf{Quality Scoring Prompt for Scenario Transcription.}}
\label{tab:prompt-quality-scoring-transcription}
\begin{tabular}{p{0.95\linewidth}}
\toprule
\textbf{System Persona} \\ \midrule
You are an expert evaluator in autonomous driving scenario creation.  
You will be given (1) the scenario-transcription prompt used to generate a textual driving scene,  
and (2) the model output produced by that prompt.

Your task is to assign a score in \texttt{[0,1]} evaluating:

1. \textbf{Faithfulness}: whether the scene description accurately reflects the traffic rule constraints included in the prompt.  

2. \textbf{Coherence}: whether the scene is internally consistent (actors, road type, environment).  

3. \textbf{Relevance}: whether the scene directly corresponds to the intended rule semantics without omitting or fabricating conditions.

Output \textbf{only a floating-point score} in \texttt{[0,1]}.

\quad\texttt{<score>}
\\
\midrule
\textbf{User Input} \\ \midrule
Scenario Transcription Prompt \\
Scenario Transcription Output \\

Output Score:

\\
\bottomrule
\end{tabular}
\end{table*}

\begin{table*}
\centering
\footnotesize
\caption{\textbf{Quality Scoring Prompt for DSL Translation.}}
\label{tab:prompt-quality-scoring-dsl}
\begin{tabular}{p{0.95\linewidth}}
\toprule
\textbf{System Persona} \\ \midrule
You are an expert evaluator for structured scenario representation and DSL authoring.  
You will be given (1) the DSL-translation prompt used to generate a YAML/DSL scenario,  
and (2) the model output produced by that prompt.

Your task is to assign a score in \texttt{[0,1]} evaluating:

1. \textbf{Structural correctness}: whether the DSL syntax follows the schema implied by the prompt.  

2. \textbf{Semantic alignment}: whether entities, relations, road network, and environment accurately reflect the scenario description in the prompt.

3. \textbf{Executability}: whether the DSL can be reliably executed in CARLA without contradictions.

Output \textbf{only a floating-point score} in \texttt{[0,1]}.

\quad\texttt{<score>}
\\
\midrule
\textbf{User Input} \\ \midrule
DSL Translation Prompt \\
DSL Translation Output \\
Output Score:

\\
\bottomrule
\end{tabular}
\end{table*}

\begin{table*}
\centering
\footnotesize
\caption{\textbf{Prompt for Testing DriveCombo.}}
\label{tab:prompt-test}
\begin{tabular}{p{0.95\linewidth}}
\toprule
\textbf{System Persona} \\ \midrule
You are a driver assistant. Your task is to answer the question based on the scenario description and question stem. 

The scenario description will be provided in the image. 

Please answer the question based on the scenario description shown in the image and question stem. 

Please return the answer in the format of "A", "B", "C", or "D". And give the reason for your answer. 

\\
\midrule
\textbf{User Input} \\ \midrule
Scenario Description: \{scenario\_description image\}

Question Stem: \{question\_stem\}

Options: \{options\}

Output:
\\
\bottomrule
\end{tabular}
\end{table*}

\begin{table*}
\centering
\footnotesize
\caption{\textbf{Prompt for Testing DriveCombo-Text variant.}}
\label{tab:prompt-test-tqa}
\begin{tabular}{p{0.95\linewidth}}
\toprule
\textbf{System Persona} \\ \midrule
You are a driver assistant. Your task is to answer the question based on the scenario description and question stem.

Please answer the question based on the scenario description and question stem.

Please return the answer in the format of "A", "B", "C", or "D". And give the reason for your answer.
\\
\midrule
\textbf{User Input} \\ \midrule
Scenario Description: \{scenario\_description text\}

Question Stem: \{question\_stem\}

Options: \{options\}

Output:
\\
\bottomrule
\end{tabular}
\end{table*}


\end{document}